\documentclass[runningheads]{llncs}
\usepackage{graphicx}

\usepackage[utf8]{inputenc} 
\usepackage[T1]{fontenc}    
\usepackage{hyperref}       
\usepackage{url}            
\usepackage{booktabs}       
\usepackage{amsfonts}       
\usepackage{nicefrac}       
\usepackage{microtype}      
\usepackage{amsmath, amssymb, amsfonts,graphicx}
\usepackage{mathtools}
\usepackage[linesnumbered,ruled,vlined]{algorithm2e}
\usepackage[noend]{algpseudocode}
\usepackage{thmtools} 

\usepackage[inline]{enumitem}
\usepackage{dsfont}

\usepackage{subcaption}

\usepackage[dvipsnames]{xcolor}

\newif \ifcomments
\commentstrue
\commentsfalse

\ifcomments
\newcommand{\MT}[1]{{\textcolor{OliveGreen}{MT: #1}}}
\newcommand{\CB}[1]{{\textcolor{blue}{CB: #1}}}
\else
\newcommand{\MT}[1]{}
\newcommand{\CB}[1]{}
\fi

\newcommand{\be}{\begin{equation}}
\newcommand{\ee}{\end{equation}}
\newcommand{\coreset}{\textsc{Coreset}}
\newcommand{\streamCore}{\textsc{Streaming-Coreset}}

\newcommand{\C}{\mathcal{C}}

\newcommand{\RD}{\REAL^{d + 1}}

\DeclareMathOperator*{\E}{\mathbb{E}}

\DeclarePairedDelimiter{\ceil}{\lceil}{\rceil}

\newcommand{\norm}[1]{\left\| #1\right\|}                               %
\newcommand{\br}[1]{\left\{#1\right\}}

\newcommand{\abs}[1]        {\left| #1\right|}

\newcommand{\eps}{\ensuremath{\varepsilon}}                       
\renewcommand{\epsilon}{\varepsilon}
\newcommand{\REAL}{\ensuremath{\mathbb{R}}}                       

\newcommand\tab[1][0.25cm]{\hspace*{#1}}

\newcommand{\PP}{\mathcal{P}}

\renewcommand\SS{\mathcal{S}}
\newcommand\BigO{\mathcal{O}}
\newcommand\Bigo{\BigO}

\newcommand{\OPT}{\widetilde{opt}}

\DeclareMathOperator*{\argmin}{\arg\!\min}
\DeclareMathOperator*{\argmax}{\arg\!\max}

\DeclarePairedDelimiterX{\dotp}[2]{\langle}{\rangle}{#1, #2}

\newcommand{\ramp}[1]{\left [ #1 \right]_+}

\newcommand{\ap}[1][y]{\alpha_{#1}^{(i)}}
\newcommand{\F}[1][\hat w]{F_\lambda(\PP, #1)}
\newcommand{\f}[1][w]{f_\lambda(p, #1)}

\newcommand{\U}[1][P]{\ensuremath{U{\left( #1 \right)}}}
\newcommand{\x}{\tilde x}

\newcommand{\grad}[1][\hat w]{\norm{\nabla \F}_2}
\newcommand{\w}[1][\hat w]{\norm{{#1}_{1:d}}_2}

\newcommand{\alphaDefNew}{\frac{\U[P \setminus P_y^{(i)}]}{2 \lambda \U \U[P_y^{(i)}]}}

\newcommand{\Dim}{\REAL^{d+1}}

\newcommand{\SamplesNeeded}{c \left(\frac{t}{\epsilon^2} \big(d \log t + \log (1 / \delta) \big) \right)}

\newcommand{\SamplesNeededO}{\Omega \left(\frac{t}{\epsilon^2} \big(d \log t + \log (1 / \delta) \big) \right)}

\newcommand{\SamplesNeededStream}{c \left(\frac{t\log^2{(n)}}{\epsilon^2} \big(d \log t + \log (\log{n} / \delta) \big) \right)}

\makeatletter
\newcommand{\printfnsymbol}[1]{%
  \textsuperscript{\@fnsymbol{#1}}%
}
\makeatother

\usepackage{diagbox}

\begin{document}
\title{On Coresets for Support Vector Machines\thanks{This research was supported in part by the U.S. National Science Foundation (NSF) under Awards 1723943 and 1526815, Office of Naval Research (ONR) Grant N00014-18-1-2830, Microsoft, and JP Morgan Chase.}}
%
%
\author{Murad Tukan\inst{1}\thanks{These authors contributed equally to this work.}
,
Cenk Baykal\inst{2}\printfnsymbol{2}, 
Dan Feldman\inst{1}
\and
Daniela Rus\inst{2}
}
\authorrunning{M. Tukan, C. Baykal, et al.}
%
\institute{University of Haifa, Computer Science Department, Israel \email{muradtuk@gmail.com, dannyf.post@gmail.com} \and
MIT CSAIL, Cambridge, USA
\email{\{baykal, rus\}@mit.edu}}
\maketitle              

\begin{abstract}
We present an efficient coreset construction algorithm for large-scale Support Vector Machine (SVM) training in Big Data and streaming applications. A coreset is a small, representative subset of the original data points such that a models trained on the coreset are provably competitive with those trained on the original data set. Since the size of the coreset is generally much smaller than the original set, our preprocess-then-train scheme has potential to lead to significant speedups when training SVM models. We prove lower and upper bounds on the size of the coreset required to obtain small data summaries for the SVM problem. As a corollary, we show that our algorithm can be used to extend the applicability of any off-the-shelf SVM solver to streaming, distributed, and dynamic data settings. We evaluate the performance of our algorithm on real-world and synthetic data sets. Our experimental results reaffirm the favorable theoretical properties of our algorithm and demonstrate its practical effectiveness in accelerating SVM training.
\end{abstract}

\section{Introduction}
\label{sec:Introduction}
Popular machine learning algorithms are computationally expensive, or worse yet, intractable to train on massive data sets, where the input data set is so large that it may not be possible to process all the data at one time. A natural approach to achieve scalability when faced with Big Data is to first conduct a preprocessing step to summarize the input data points by a significantly smaller, representative set. Off-the-shelf training algorithms can then be run efficiently on this compressed set of data points. The premise of this two-step learning procedure is that the model trained on the compressed set will be provably competitive with the model trained on the original set -- as long as the data summary, i.e., the \emph{coreset}, can be generated efficiently and is sufficiently representative.

Coresets are small weighted subsets of the training points such that models trained on the coreset are approximately as good as the ones trained on the original (massive) data set.
Coreset constructions were originally introduced in the context of computational geometry~\cite{agarwal2005geometric} and subsequently generalized for applications to other problems, such as logistic regression, neural network compression, and mixture model training~\cite{baykal2018data,braverman2016new,feldman2011unified,langberg2010universal,liebenwein2019provable} (see~\cite{feldman2019core} for a survey).

A popular coreset construction technique -- and the one that we leverage in this paper -- is to use importance sampling with respect to the points' \emph{sensitivities}. The sensitivity of each point is defined to be the worst-case relative impact of each data point on the objective function. Points with high sensitivities have a large impact on the objective value and are sampled with correspondingly high probability, and vice-versa. 
The main challenge in generating small-sized coresets often lies in evaluating the importance of each point in an accurate and computationally-efficient way.

\subsection{Our Contributions}
In this paper, we propose an efficient coreset construction algorithm to generate compact representations of large data sets to accelerate SVM training. Our approach hinges on bridging the SVM problem with that of $k$-means clustering. 
As a corollary to our theoretical analysis, we obtain theoretical justification for the widely reported empirical success of using $k$-means clustering as a way to generate data summaries for large-scale SVM training. In contrast to prior approaches, our approach is both (i) provably efficient and (ii) naturally extends to streaming or dynamic data settings. Above all, our approach can be used \emph{to enable the applicability of any off-the-shelf SVM solver} -- including gradient-based and/or approximate ones, e.g., Pegasos~\cite{shalev2011pegasos}, to streaming and distributed data settings by exploiting the \emph{composibility} and \emph{reducibility} properties of coresets~\cite{feldman2019core}.

In particular, this paper contributes the following:
\begin{enumerate}
	\item A coreset construction algorithm for accelerating SVM training based on an efficient importance sampling scheme.
    \item An analysis proving lower bounds on the number of samples required by any coreset construction algorithm to approximate the input data set.
    \item Theoretical guarantees on the efficiency and accuracy of our coreset construction algorithm.
    \item Evaluations on synthetic and real-world data sets that demonstrate the effectiveness of our algorithm in both streaming and offline settings.
\end{enumerate}
\section{Related Work}
\label{sec:RelatedWork}
Training SVMs requires $\BigO(n^3)$ time and $\BigO(n^2)$ space in the offline setting where $n$ is the number of training points. Towards the goal of accelerating SVM training in the offline setting, \cite{tsang2007simpler,tsang2005core} introduced the Core Vector Machine (CVM) and Ball Vector Machine (BVM) algorithms, which are based on reformulating the SVM problem as the Minimum Enclosing Ball (MEB) problem and Enclosing Ball (EB) problem, respectively, and by leveraging existing coreset constructions for each; see~\cite{badoiu2003smaller}. However, CVM's accuracy and convergence properties have been noted to be at times inferior relative to those of existing SVM implementations~\cite{loosli2007comments}; moreover, unlike the algorithm presented in this paper, neither the CVM, nor the BVM algorithm extends naturally to streaming or dynamic settings where data points are continuously inserted or deleted. Similar geometric approaches, including extensions of the MEB formulation, those based on convex hulls and extreme points, among others, were investigated by \cite{agarwal2010streaming,gartner2009coresets,har2007maximum,joachims2006training,nandan2014fast,rai2009streamed}. Another class of related work includes the use of canonical optimization algorithms such as the Frank-Wolfe algorithm~\cite{clarkson2010coresets}, Gilbert's algorithm~\cite{clarkson2010coresets,clarkson2012sublinear}, and a primal-dual approach combined with Stochastic Gradient Descent (SGD)~\cite{hazan2011beating}. 

SGD-based approaches, such as Pegasos~\cite{shalev2011pegasos}, have been a popular tool of choice in approximately-optimal SVM training. Pegasos is a stochastic sub-gradient algorithm for obtaining a $(1+\epsilon)$-approximate solution to the SVM problem in $\widetilde{\Bigo}(dn\lambda/\epsilon)$ time for a linear kernel, where $\lambda$ is the regularization parameter and $d$ is the dimensionality of the input data points. In contrast to our method, these approaches and their corresponding theoretical guarantees do not feasibly extend to dynamic data sets and/or streaming settings.
In particular, gradient-based approaches cannot be trivially extended to streaming settings since the arrival of each input point in the stream results in a change of the gradient.

There has been prior work in streaming algorithms for SVMs, such as those of~\cite{agarwal2010streaming,har2007maximum,nathan2014accurate,rai2009streamed}. However, these works generally suffer from poor practical performance in comparison to that of approximately optimal SVM algorithms in the offline (batch) setting, high difficulty of implementation and application to practical settings, and/or lack of strong theoretical guarantees. Unlike the algorithms of prior work, our method is simultaneously simple-to-implement, exhibits theoretical guarantees, and naturally extends to streaming and dynamic data settings, where the input data set is so large that it may not be possible to store or process all the data at one time. 

\section{Problem Definition}
\label{sec:problem-definition}
Let $P = \br{ (x, y) \, : \, x \in \REAL^d \times {1}, y \in \br{\pm 1}}$ denote a set of $n$ input points. Note that for each point $p = (x,y) \in P$, the last entry $x_{d+1} = 1$ of $x$ accounts for the bias term  embedding into the feature space\footnote{We perform this embedding for ease of presentation later on in our analysis.}. To present our results with full generality, we consider the setting where the input points $P$ may have weights associated with them. Hence, given $P$ and a weight function $u: P \to \mathbb{R}_{\ge 0}$, we let $\PP = (P, u)$ denote the weighted set with respect to $P$ and $u$. The canonical unweighted case can be represented by the weight function that assigns a uniform weight of 1 to each point, i.e.,  $u(p) = 1$ for every point $p \in P$. For every $T \subseteq P$, let $\U[T] = \sum_{p \in T} u(p)$. We consider the scenario where $n$ is much larger than the dimension of the data points, i.e., $n \gg d$.


For a normal to a separating hyperplane $w \in \mathbb{R}^{d+1}$, let $w_{1:d}$ denote vector which contains the first $d$ entries of $w$. The last entry of $w$ ($w_{d+1}$) encodes the bias term $b \in \REAL$. Under this setting, the hinge loss of any point $p = (x, y) \in P$ with respect to a normal to a separating hyperplane, $w \in \REAL^{d+1}$, is defined as $h(p, w) = \ramp{1 - y \dotp{x}{w}}$, where $\ramp{\cdot} = \max \{0, \cdot\}$. As a prelude to our subsequent analysis of sensitivity-based sampling, we quantify the contribution of each point $p = (x,y) \in P$ to the SVM objective function as
\begin{equation}
\f = \frac{1}{2\U} \norm{w_{1:d}}_2^2 + \lambda h(p,w),
\end{equation}
where $\lambda \in [0,1]$ is the SVM regularization parameter, and
$
h(p,w) = \ramp{1 - y \dotp{x}{w}}
$
is the hinge loss with respect to the query $w \in \REAL^{d+1}$ and point $p = (x,y)$. Putting it all together, we formalize the $\lambda$-regularized SVM problem as follows.
\begin{definition}[$\lambda$-regularized SVM Problem]
\label{def:svm-problem}
For a given weighted set of points $\PP = (P, u)$ and a regularization parameter $\lambda \in [0,1]$, the $\lambda$-regularized SVM problem with respect to $\PP$ is given by
\[
\min_{w \in \REAL^{d+1}} F_\lambda(\PP,w),
\]
where 
\begin{equation}
\label{eqn:svm-obj}
F_\lambda(\PP,w) = \sum\limits_{p \in \PP} u(p) f(p,w).
\end{equation}
\end{definition}
We let $w^*$ denote the optimal solution to the SVM problem with respect to $\PP$, i.e., $w^* \in \argmin_{w \in \mathbb{R}^{d+1}} F_\lambda(\PP, w)$. A solution $\hat{w} \in \Dim$ is an $\xi$-approximation to the SVM problem if $F_\lambda(\PP, \hat w) \leq F_\lambda(\PP,w^*) + \xi$. Next, we formalize the \emph{coreset guarantee} that we will strive for when constructing our data summaries.

\paragraph{Coresets.}
A coreset is a compact representation of the full data set that provably approximates the SVM cost function \eqref{eqn:svm-obj} for \emph{every query} $w \in \REAL^{d+1}$ -- including that of the optimal solution $w^*$. We formalize this notion below for the SVM problem with objective function $F_\lambda(\cdot)$ as in \eqref{eqn:svm-obj} below.
\begin{definition}[$\eps$-coreset]
\label{def:epsCoreset}
Let $\epsilon \in (0,1)$ and let $\PP = (P, u)$ be the weighted set of training points as before. A weighted subset $\SS = (S,v)$, where $S \subset P$ and $v: S \to \mathbb{R}_{\ge 0}$ is an $\epsilon$-coreset for $\PP$ if
\begin{align}
\label{eqn:coreset-property}
\forall{w \in \REAL^{d+1}} \quad 
\left|F_\lambda\left(\PP,w\right)-F_\lambda\left(\SS, w \right)\right| \leq \eps F_\lambda\left(\PP,w\right).
\end{align}
\end{definition}
This strong guarantee implies that the models trained on the coreset $\SS$ with \emph{any} off-the-shelf SVM solver will be approximately (and provably) as good as the optimal solution $w^*$ obtained by training on the entire data set $\PP$. This also implies that, if the size of the coreset is provably small, e.g., logartihmic in $n$ (see Sec.~\ref{sec:analysis}), then an approximately optimal solution can be obtained much more quickly by training on $\SS$ rather than $\PP$, leading to  computational gains in practice for both offline and streaming data settings (see Sec.~\ref{sec:results}). 





The difficulty in constructing coresets lies in constructing them (i) \emph{efficiently}, so that the preprocess-then-train pipeline takes less time than training on the full data set and (ii) \emph{accurately}, so that important data points -- i.e., those that are imperative to obtaining accurate models -- are not left out of the coreset, and redundant points are eliminated so that the coreset size is small. In the following sections, we introduce and analyze our coreset algorithm for the SVM problem.

\section{Method}
\label{sec:method}
Our coreset construction scheme is based on the unified framework of~\cite{feldman2011unified,langberg2010universal} and is shown in Alg.~\ref{algorithm}. The crux of our algorithm lies in generating the importance sampling distribution via efficiently computable upper bounds (proved in Sec.~\ref{sec:analysis}) on the importance of each point (Lines~\ref{alg:L1}--\ref{alg:sens_bound}). Sufficiently many points are then sampled from this distribution and each point is given a weight that is inversely proportional to its sample probability (Lines~\ref{alg:sum_sens}--\ref{alg:sample}). The number of points required to generate an $\epsilon$-coreset with probability at least $1 - \delta$ is a function of the desired accuracy $\epsilon$, failure probability $\delta$, and complexity of the data set ($t$ from Theorem~\ref{thm:epsilon-coreset}). Under mild assumptions on the problem at hand (see Sec.~\ref{par:sufficient}), the required sample size is polylogarithmic in $n$.

\begin{algorithm}[!htb]
\SetKwInOut{Input}{Input}
\SetKwInOut{Output}{Output}
\DontPrintSemicolon
\Input{A set of training points $P \subseteq \mathbb{R}^{d+1} \times \{-1,1\}$ containing $n$ points, weight function $u : P \to \REAL_{\geq 0}$, a regularization parameter $\lambda \in [0,1]$, an approximation factor $\xi > 0$, a positive integer $k$, a sample size $m$}
\Output{An weighted set $(S,v)$ which satisfies Theorem~\ref{thm:epsilon-coreset}} 
\caption{$\coreset(P,u,\lambda,\xi,k,m)$\label{one}}
\label{algorithm}
\vskip -0.02in
$\Tilde{w} \gets$ An $\xi$-approximation for the optimal SVM of $(P,u)$\label{alg:L1}; \\
$\OPT_\xi \gets \F[\Tilde{w}] - \xi$; \label{alg:L2}

\For{$y \in \{-, +\}$\label{alg:start_for}}{
    $P_y \gets $ all the points in $P$ that are associated with the label $y$; \label{alg:L4} \\
    $\left(c_y^{(i)}, P_y^{(i)} \right)_{i=1}^k \gets $ \textsc{k-means++}($\PP$, $k$); \label{alg:L5}\\
    \For{every $i \in [k]$}{
        $\ap \gets \frac{\U[P \setminus P_y^{(i)}]}{2\lambda \U \U[P_y^{(i)}]}$; \label{alg:alpha}\\
        \For{every $p = (x,y) \in P_y^{(i)}$}{
            $p_\Delta \gets c_y^{(i)} - yx$; \label{alg:L8}\\
            $\gamma(p) \gets \frac{u(p)}{\U[P_y^{(i)}]} + \lambda u(p) \frac{9}{2}\max\br{\frac{4}{9}\ap, \sqrt{4\left(\ap\right)^2 + \frac{2\norm{p_\Delta}_2^2}{9 \OPT_\xi} } - 2\ap }$; \label{alg:sens_bound}
        }
    }
    
}

$t \gets \sum_{p \in P} \gamma(p)$; \label{alg:sum_sens} \\
$(S, v) \gets m$ weighted samples from $\PP = (P,u)$ where each point $p \in P$ is sampled with probability $q(p) = \frac{\gamma(p)}{t}$ and, if sampled, has weight $v(p) = \frac{u(p)}{m q(p)}$; \label{alg:sample}\\
 \Return $(S,v)$; \label{alg:L11}\\
\end{algorithm}

Our algorithm is an importance sampling procedure that first generates a judicious sampling distribution based on the structure of the input points and samples sufficiently many points from the original data set. The resulting weighted set of points $\SS = (S,v)$, serves as an unbiased estimator for $\F[w]$ for any query $w \in \REAL^{d+1}$, i.e., $\E[ F_\lambda\left(\SS, w\right)] = \F[w]$. Although sampling points uniformly with appropriate weights can also generate such an unbiased estimator, it turns out that the variance of this estimation is minimized if the points are sampled according to the distribution defined by the ratio between each point's sensitivity and the sum of sensitivities, i.e., $\gamma(p) / t$ on Line~\ref{alg:sample}~\cite{bachem2017practical}.

\subsection{Computational Complexity}
Coresets are intended to provide efficient and provable approximations to the optimal SVM solution. However, the very first line of our algorithm entails computing an (approximately) optimal solution to the SVM problem. This seemingly eerie phenomenon is explained by the merge-and-reduce technique~\cite{har2004coresets} that ensures that our coreset algorithm is only run against small partitions of the original data set \cite{braverman2016new,har2004coresets,lucic2017training}. The merge-and-reduce approach (depicted in Alg.~\ref{algorithmStreams} in Sec.~\ref{sec:streaming-extension} of the appendix) leverages the fact that coresets are composable and reduces the coreset construction problem for a (large) set of $n$ points into the problem of computing coresets for $\frac{n}{2 |S|}$ points, where $2|S|$ is the minimum size of input set that can be reduced to half using Algorithm~\ref{algorithm}~\cite{braverman2016new}. Assuming that the sufficient conditions for obtaining polylogarithmic size coresets implied by Theorem~\ref{thm:epsilon-coreset} hold, the overall time required  is approximately linear in $n$. 

\section{Analysis}
\label{sec:analysis}
In this section, we analyze the sample-efficiency and computational complexity of our algorithm. The outline of this section is as follows: we first formalize the importance (i.e., \emph{sensitivity}) of each point and summarize the necessary conditions for the existence of small coresets. We then present the negative result that, in general, sublinear coresets do not exist for \emph{every} data set (Lem.~\ref{lem:sens-lower-bound}). Despite this, we show that we can obtain accurate approximations for the sensitivity of each point via an approximate $k$-means clustering (Lems.~\ref{lem:sens-upper-bound} and~\ref{lem:sum-sens-upper-bound}), and present non-vacuous, data-dependent bounds on the sample complexity (Thm.~\ref{thm:epsilon-coreset}). Our technical results in full with corresponding proofs can be found in the Appendix. 



\subsection{Preliminaries}
We will henceforth state all of our results with respect to the weighted set of training points $\PP = (P, u)$, $\lambda \in [0,1]$, and SVM cost function $F_{\lambda}$ (as in Sec.~\ref{sec:problem-definition}). The definition below rigorously quantifies the \emph{relative contribution} of each point.
\begin{definition}[Sensitivity~\cite{braverman2016new}]
\label{def:sensitivity}
The sensitivity of each point $p \in P$ is given by
\begin{equation}
s(p) = \sup_{w} \,  \frac{u(p) \f}{\F[w]}.
\end{equation}
\end{definition}
Note that in practice, exact computation of the sensitivity is intractable, so we usually settle for (sharp) upper bounds on the sensitivity $\gamma(p) \ge s(p)$ (e.g., as in Alg.~\ref{algorithm}). Sensitivity-based importance sampling then boils down to normalizing the sensitivities by the normalization constant -- to obtain an importance sampling distribution -- which in this case is the \emph{sum of sensitivities} $t = \sum_{p \in P} s(p)$. It turns out that the required size of the coreset is at least linear in $t$~\cite{braverman2016new}, which implies that one immediate necessary condition for sublinear coresets is $t \in o(n)$.


\subsection{Lower bound for Sensitivity}
The next lemma shows that a sublinear-sized coreset cannot be constructed for \emph{every} SVM problem instance. The proof of this result is based on demonstrating a hard point set for which
the sum of sensitivities is $\Omega(n\lambda)$, ignoring $d$ factors, which implies that sensitivity-based importance sampling roughly boils down to uniform sampling for this data set. This in turn implies that if the regularization parameter is too large, e.g., $\lambda = \theta(1)$, and if $d \ll n$ (as in Big Data applications) then the required number of samples for property \eqref{eqn:coreset-property} to hold is $\Omega(n)$.

\begin{restatable}{lemma}{senslowerbound}
\label{lem:sens-lower-bound}
For an even integer $d \ge 2$, there exists a set of weighted points $\PP = (P, u)$ such that
$$
s(p) \geq \frac{n\lambda + d^2}{n \left( \lambda + d^2\right)} \qquad \forall{p \in P} \qquad \text{and} \qquad \sum\limits_{p \in P} s(p) \geq \frac{n\lambda + d^2}{\left( \lambda + d^2\right)}.
$$
\end{restatable}
We next provide upper bounds on the sensitivity of each data point with respect to the complexity of the input data. Despite the non-existence results established above, our upper bounds shed light into the class of problems for which small-sized coresets are ensured to exist.

\subsection{Sensitivity Upper Bound}
\label{subsec:sens-upper-bound}
In this subsection we present sharp, data-dependent upper bounds on the sensitivity of each point. Our approach is based on an approximate solution to the $k$-means clustering problem and to the SVM problem itself (as in Alg.~\ref{algorithm}). To this end, we will henceforth let $k$ be a positive integer, $\xi \in [0, \F[w^*]]$ be the error of the (coarse) SVM approximation, and let $(c_{y}^{(i)}, P_y^{(i)})$, $\ap$ and $p_\Delta$ for every $y \in \br{+,-}$, $i \in [k]$ and $p \in P$ as in Lines~\ref{alg:L4}--\ref{alg:L8} of Algorithm~\ref{algorithm}.

\begin{restatable}{lemma}{sensupperbound}
\label{lem:sens-upper-bound}
Let $k$ be a positive integer, $\xi \in [0, \F[w^*]]$, and let $\PP = (P,u)$ be a weighted set. Then for every $i \in [k]$, $y \in \br{+,-}$ and $p \in P_y^{(i)}$, 
\[
s(p) \leq \frac{u(p)}{\U[P_y^{(i)}]} + \lambda u(p) \frac{9}{2}\max\br{\frac{4}{9}\ap, \sqrt{4\left(\ap\right)^2 + \frac{2\norm{p_\Delta}_2^2}{9 \OPT_\xi} } - 2\ap } = \gamma(p).
\]
\end{restatable}

\begin{restatable}{lemma}{sumsensupperbound}
\label{lem:sum-sens-upper-bound}
In the context of Lemma~\ref{lem:sens-upper-bound}, the sum of sensitivities is bounded by
\[
\sum\limits_{p \in P} s(p) \leq t = 4k + \sum\limits_{i=1}^k \frac{3\lambda \text{Var}_+^{(i)}}{\sqrt{2\OPT_\xi}} + \frac{3\lambda \text{Var}_-^{(i)}}{\sqrt{2\OPT_\xi}}, 
\]
where $\text{Var}_y^{(i)} = \sum_{p \in P_y^{(i)}} u(p) \norm{p_\Delta}_2$ for all $i \in [k]$ and $y \in \br{+,-}$.
\end{restatable}

\begin{restatable}{theorem}{epsiloncoreset}
\label{thm:epsilon-coreset}
For any $\epsilon \in (0, 1/2), \delta \in (0,1)$, let $m$ be an integer satisfying
\[
m \in \SamplesNeededO,
\]
where $t$ is as in Lem.~\ref{lem:sum-sens-upper-bound}. Invoking $\coreset$ with the inputs defined in this context yields a $\epsilon$-coreset $\SS = (S, v)$ with probability at least $1 - \delta$ in $\Bigo\left( nd + T \right)$ time, where $T$ represents the computational complexity of obtaining an $\xi$-approximated solution to SVM and applying $k$-means++ on $P_+$ and $P_-$.
\end{restatable}

We refer the reader to Sec.~\ref{sec:thm-epsilon-coreset-proof} of the Appendix for the sufficient conditions required for obtaining poly-logarithmic sized coreset, and to Sec.~\ref{sec:logic-kmeans} of the Appendix for additional details on the effect of the $k$-means clustering on the sensitivity bounds.

\section{Results}
\label{sec:results}
In this section, we present experimental results that demonstrate and compare the effectiveness of our algorithm on a variety of synthetic and real-world data sets in offline and streaming data settings~\cite{Lichman:2013}. Our empirical evaluations demonstrate the practicality and wide-spread effectiveness of our approach: our algorithm consistently generated more compact and representative data summaries, and yet incurred a negligible increase in computational complexity when compared to  uniform sampling. Additional results and details of our experimental setup and evaluations can be found in Sec.~\ref{sec:exp-details} of the Appendix.
\begin{table}[htb!]
\caption{The number of input points and measurements of the total sensitivity computed empirically for each data set in the offline setting. The sum of sensitivities is significantly less than $n$ for virtually all of the data sets, which, by Thm.~\ref{thm:epsilon-coreset}, ensures the sample-efficiency of our approach on the evaluated scenarios.}  
\vspace{1em}
    \centering
    \begin{tabular}{|l|*{6}{c|}}\hline
\backslashbox[40mm]{Measurements}{Dataset}
&\makebox[3.6em]{HTRU}&\makebox[3.6em]{Credit}&\makebox[3.6em]{Pathol.}&\makebox[3.6em]{Skin}&\makebox[3.6em]{Cod}&\makebox[3.6em]{W1}\\\hline
Number of data-points ($n$) & $17,898$ & $30,000$ & $1,000$ & $245,057$ & $488,565$ & $49,749$ \\\hline
Sum of Sensitivities ($t$) & $475.8$ & $1,013.0$ & $77.6$ & $271.5$ & $2,889.2$ & $24,231.6$ \\\hline
$t / n$ (Percentage) & $2.7\%$ & $3.4\%$ & $7.7\%$ & $0.1\%$ & $0.6\%$ & $51.3\%$ \\\hline
\end{tabular}
    \label{table}
    \vspace{-2em}
\end{table}
\paragraph{Evaluation} We considered $6$ real-world data sets of varying size and complexity as depicted in Table~\ref{table} (also see Sec.~\ref{sec:exp-details} of the Appendix). For each data set of size $n$, we selected a set of $M = 15$ geometrically-spaced subsample sizes $m_1,\ldots,m_\text{M} \subset [\log{n}, n^{4/5}]$. For each sample size $m$, we ran each algorithm (Alg.~\ref{algorithm} or uniform sampling) to construct a subset $\SS = (S, v)$ of size $m$. We then trained the SVM model as per usual on this subset to obtain an optimal solution with respect to the coreset $\SS$, i.e., $w_{\SS}^* = \argmin_{w} F_\lambda(\SS, w)$. We then computed the relative error incurred by the solution computed on the coreset ($w_\SS^*$) with respect to the ground-truth optimal solution computed on the entire data set ($w^*$): 
$
\nicefrac{\abs{F_\lambda(P, w_{\SS}^*) - F_\lambda(P, w^*)}}{F_\lambda(P, w^*)};
$
See Corollary~\ref{cor:error-bound} at Sec.~\ref{sec:thm-epsilon-coreset-proof} of the Appendix.
The results were averaged across $100$ trials.
\begin{figure*}[htb!]
  \centering
  \begin{minipage}[b]{0.32\textwidth}
  \centering
 \includegraphics[width=1\textwidth]{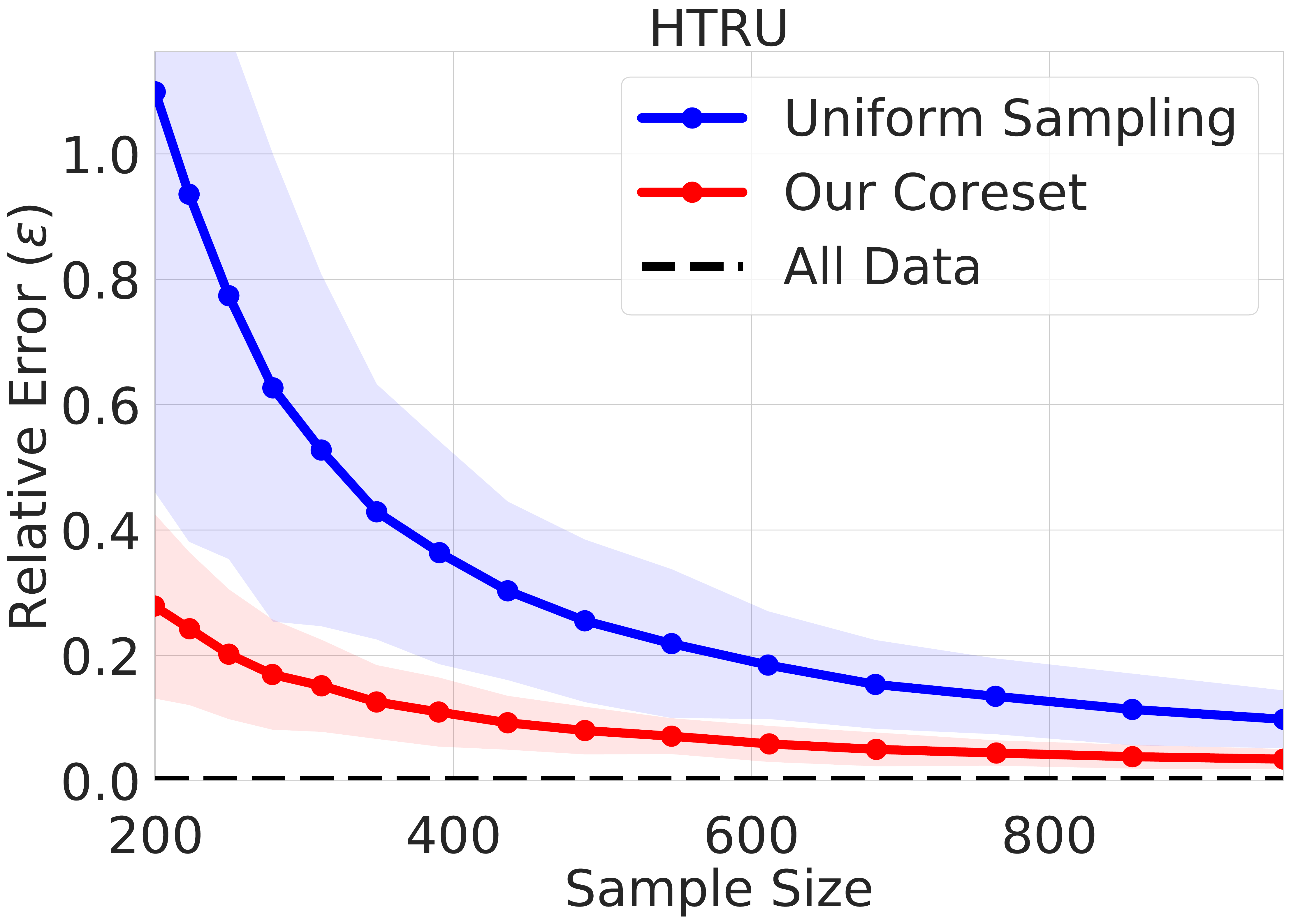}
  \end{minipage}%
  \begin{minipage}[b]{0.32\textwidth}
  \centering
 \includegraphics[width=1\textwidth]{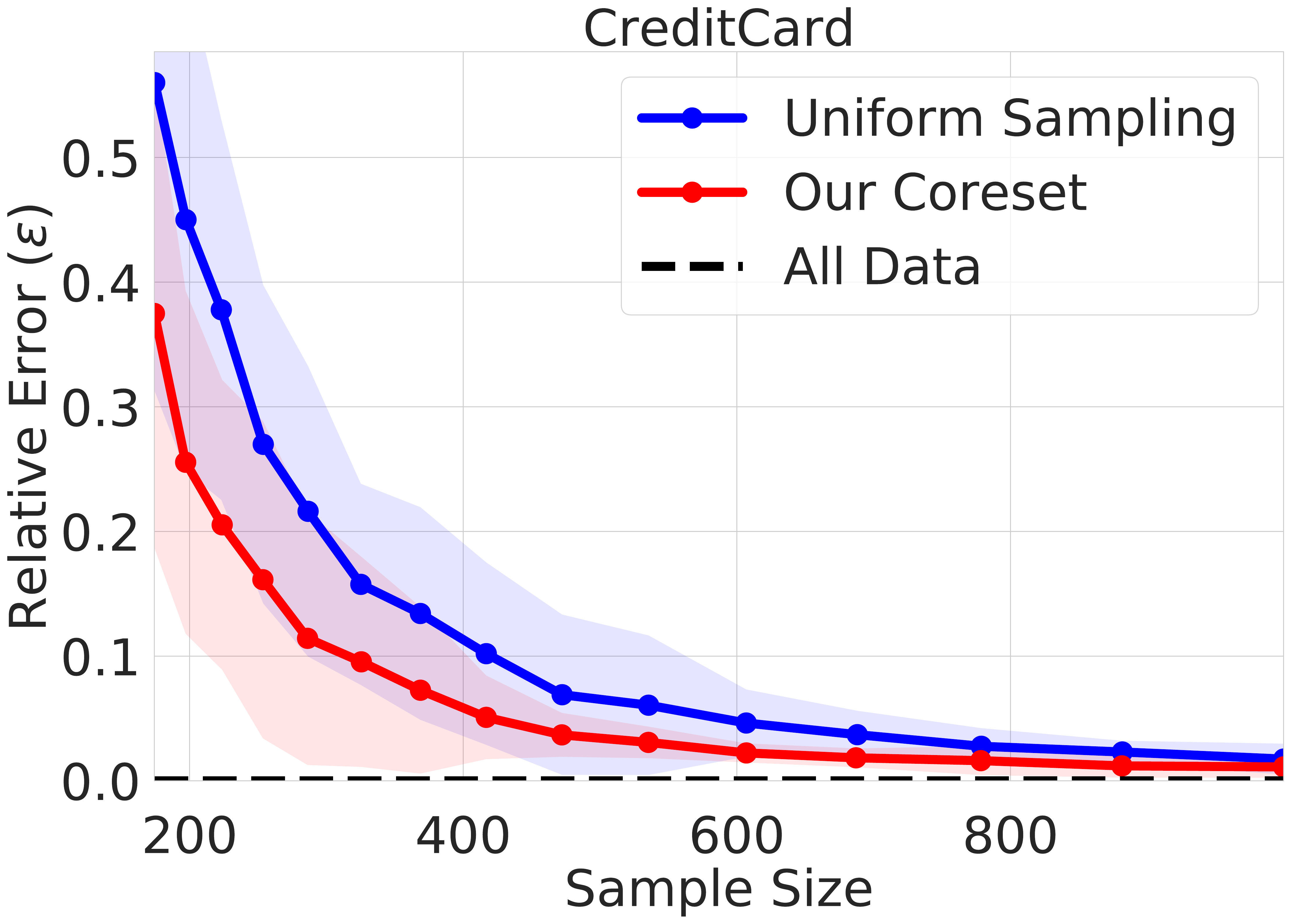}
  \end{minipage}%
  \begin{minipage}[b]{0.32\textwidth}
  \centering
 \includegraphics[width=1\textwidth]{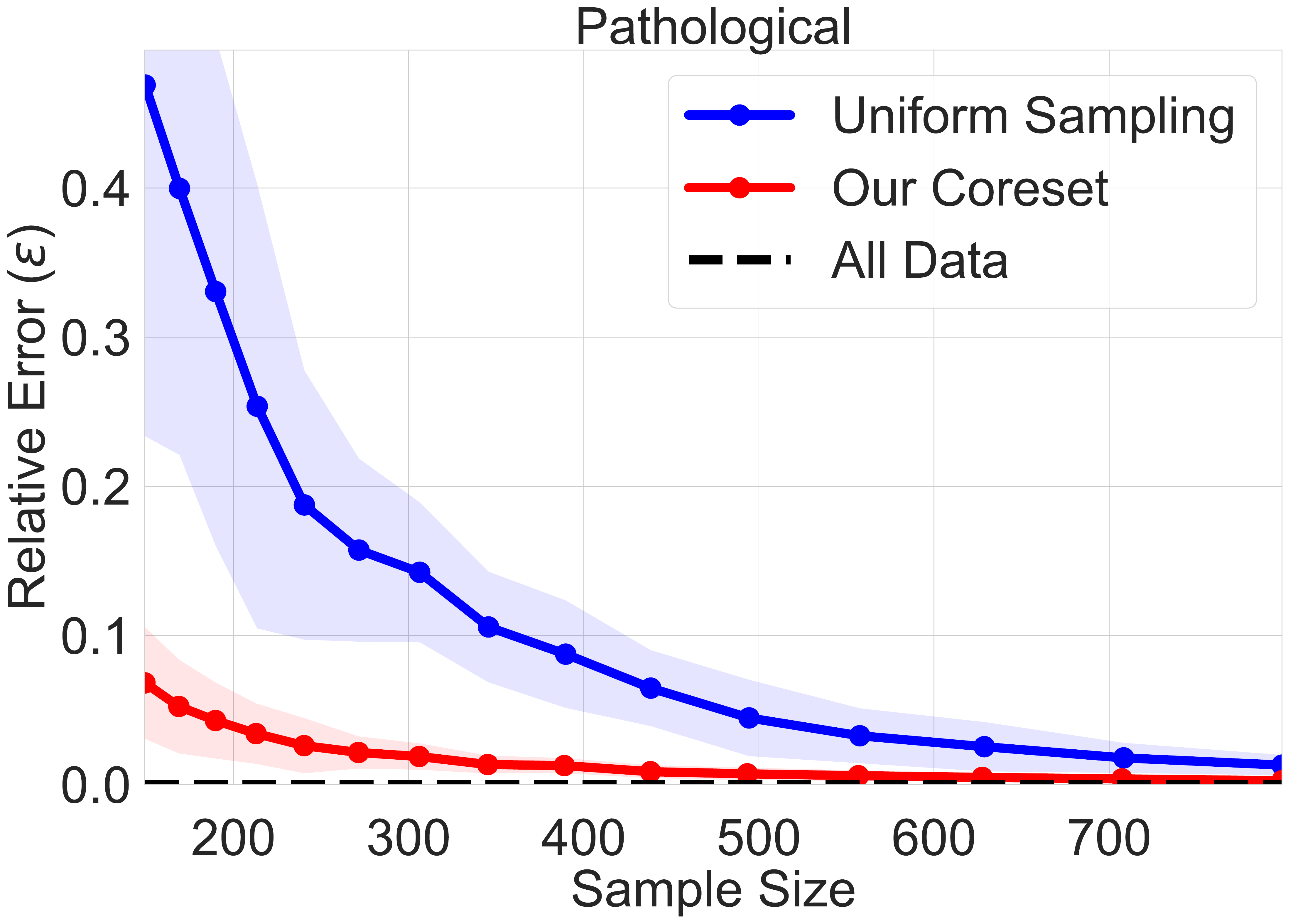}
  \end{minipage}%
  
\begin{minipage}[b]{0.32\textwidth}
  \centering
 \includegraphics[width=1\textwidth]{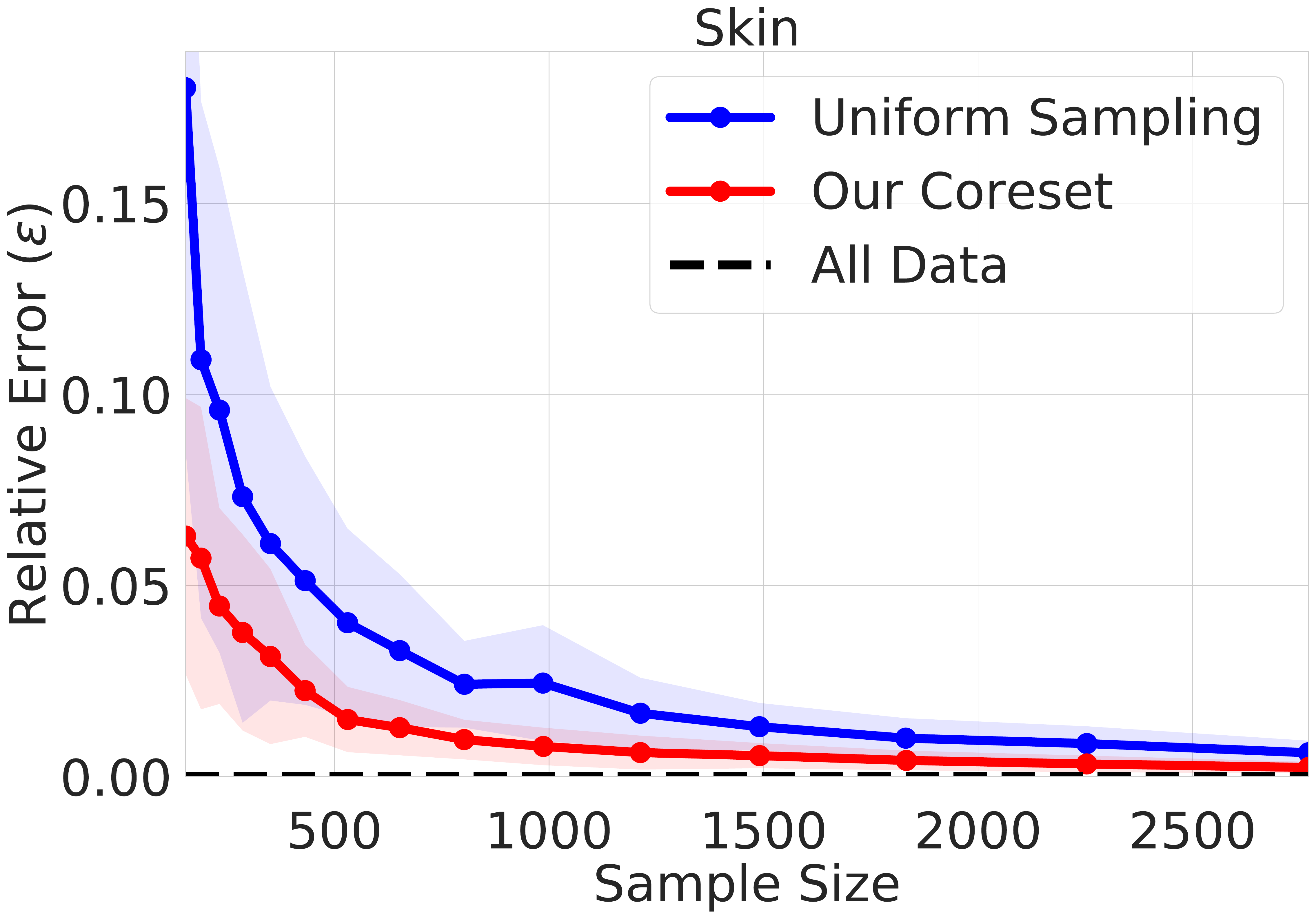}
  \end{minipage}%
  \begin{minipage}[b]{0.32\textwidth}
  \centering
 \includegraphics[width=1\textwidth]{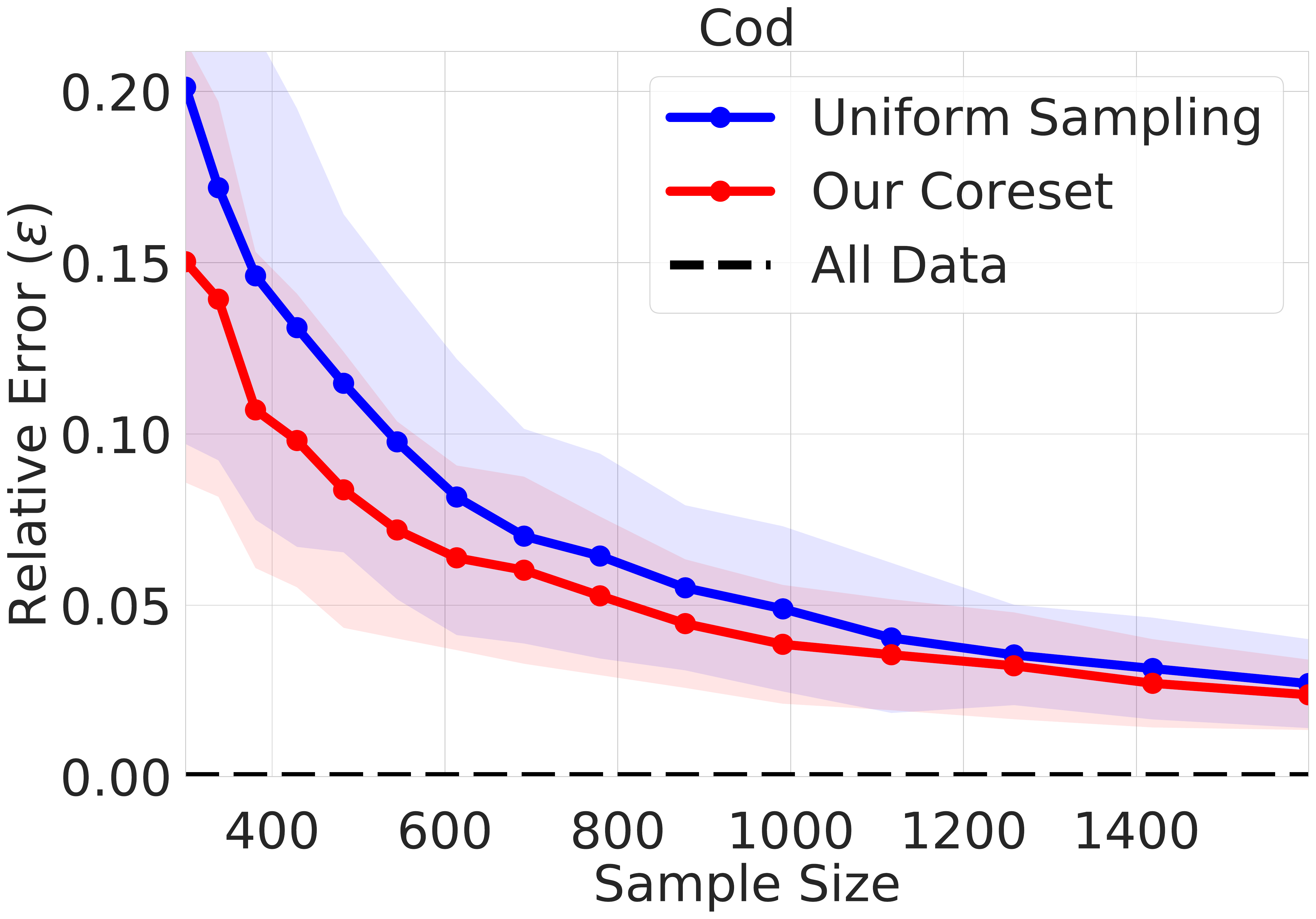} 
   \end{minipage}%
  \begin{minipage}[b]{0.32\textwidth}
  \centering
 \includegraphics[width=1\textwidth]{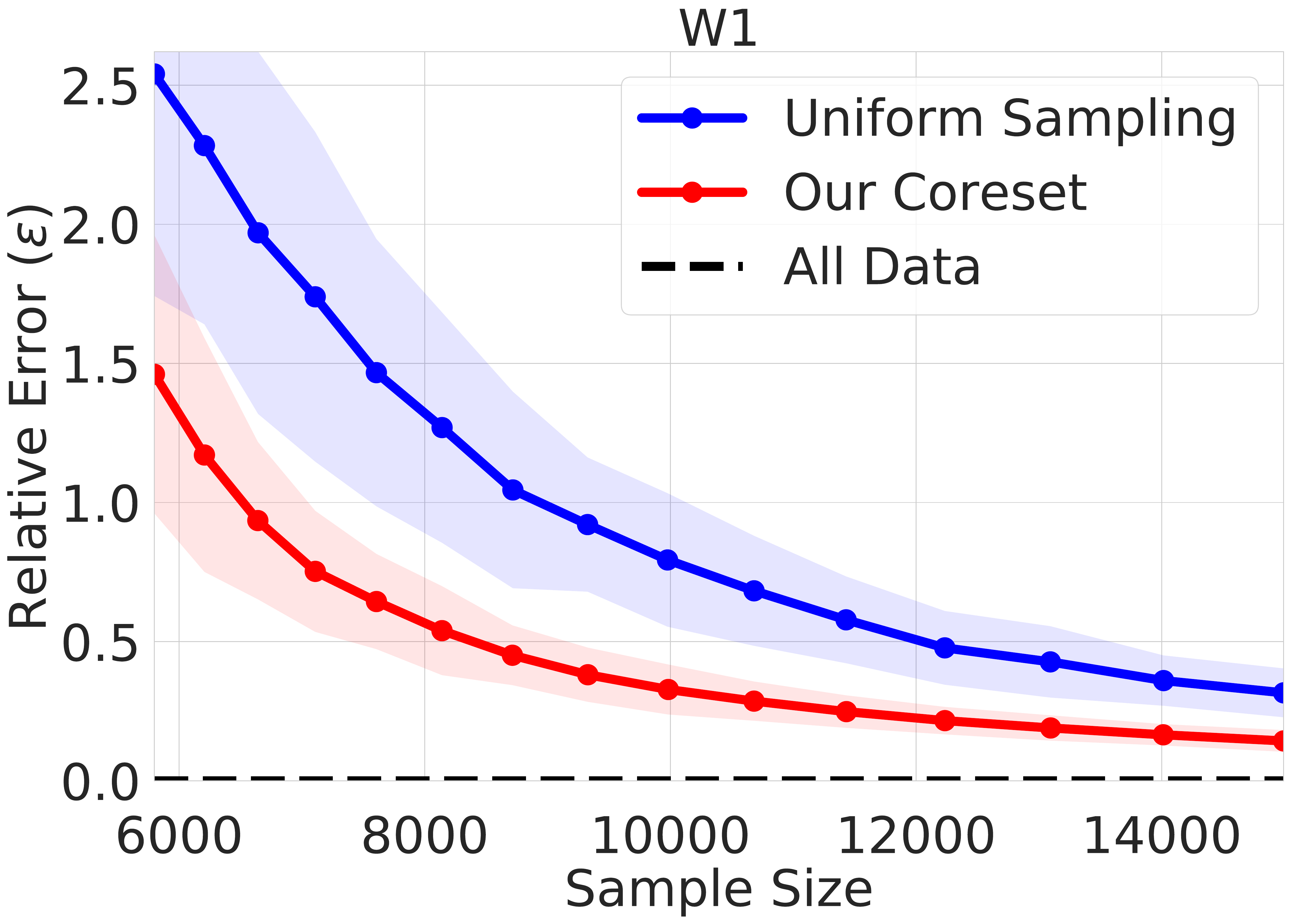}
  \end{minipage}%
		\caption{The relative error of query evaluations with respect uniform and coreset subsamples for the 6 data sets in the offline setting. Shaded region corresponds to values within one standard deviation of the mean.} 
	\label{fig:RelativeError}
\end{figure*}

\begin{figure*}[!htb]
  \centering
  \begin{minipage}[b]{0.32\textwidth}
  \centering
 \includegraphics[width=1\textwidth]{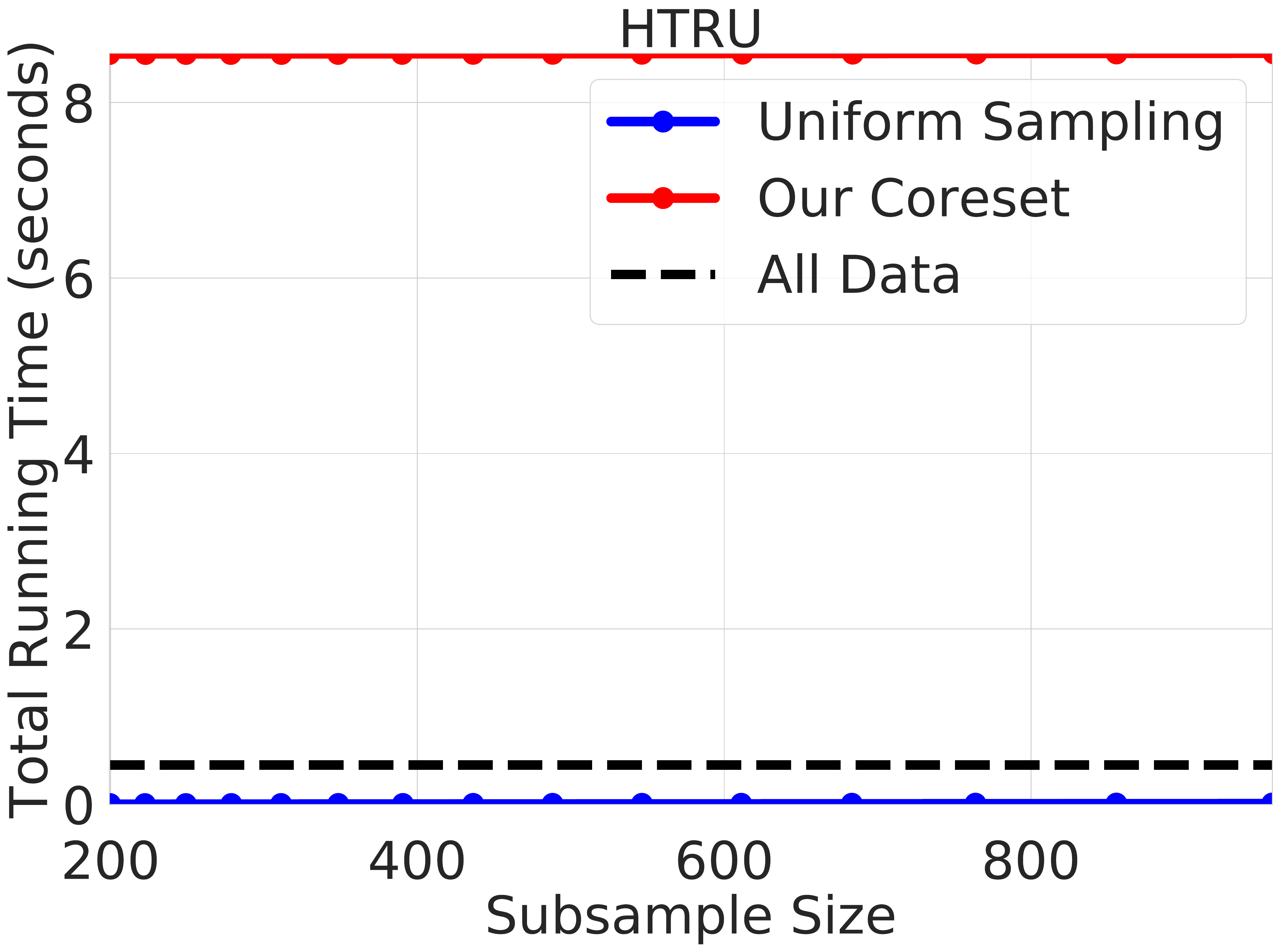}
  \end{minipage}%
  \begin{minipage}[b]{0.32\textwidth}
  \centering
 \includegraphics[width=1\textwidth]{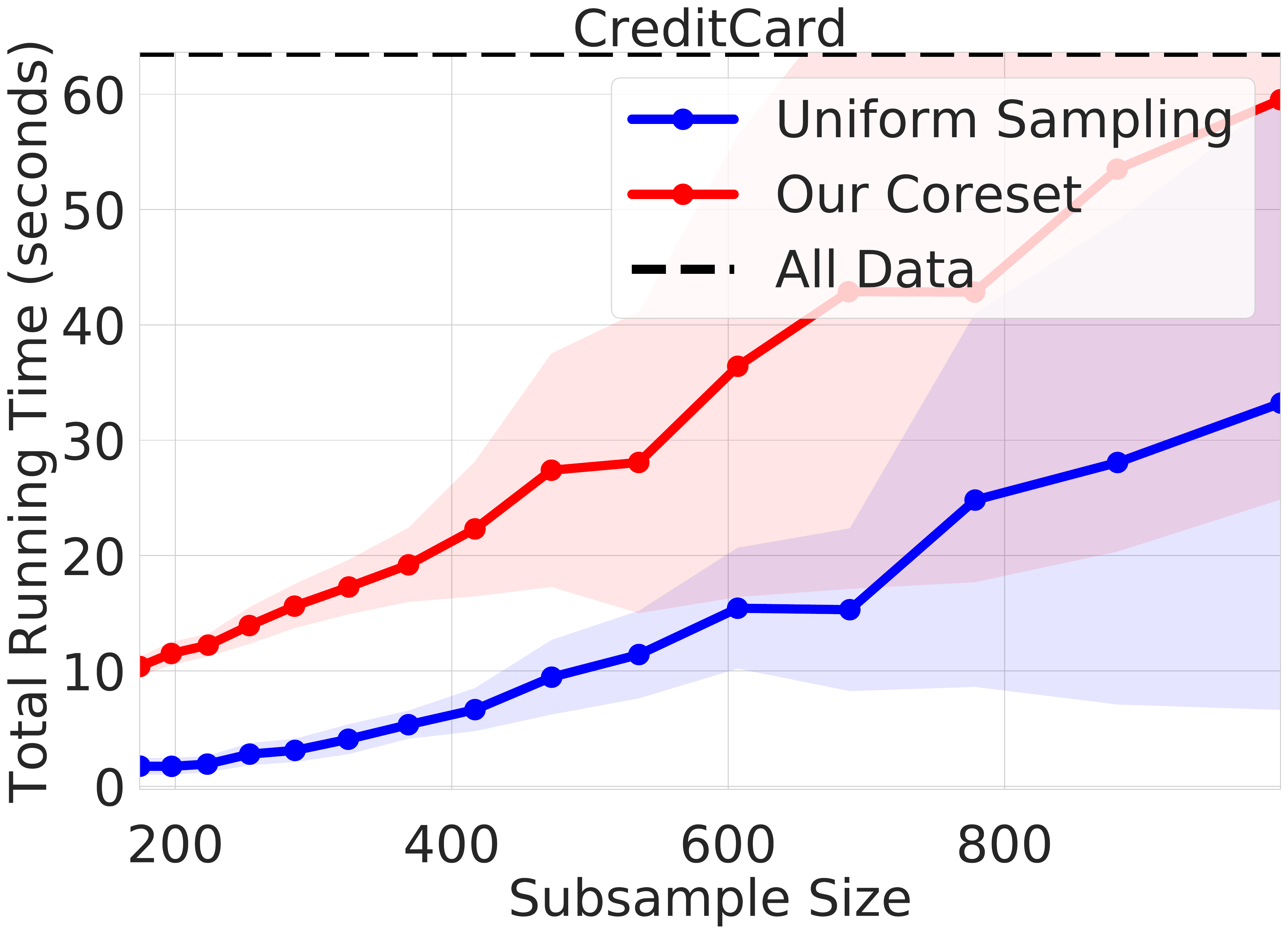}
  \end{minipage}%
  \begin{minipage}[b]{0.32\textwidth}
  \centering
 \includegraphics[width=1\textwidth]{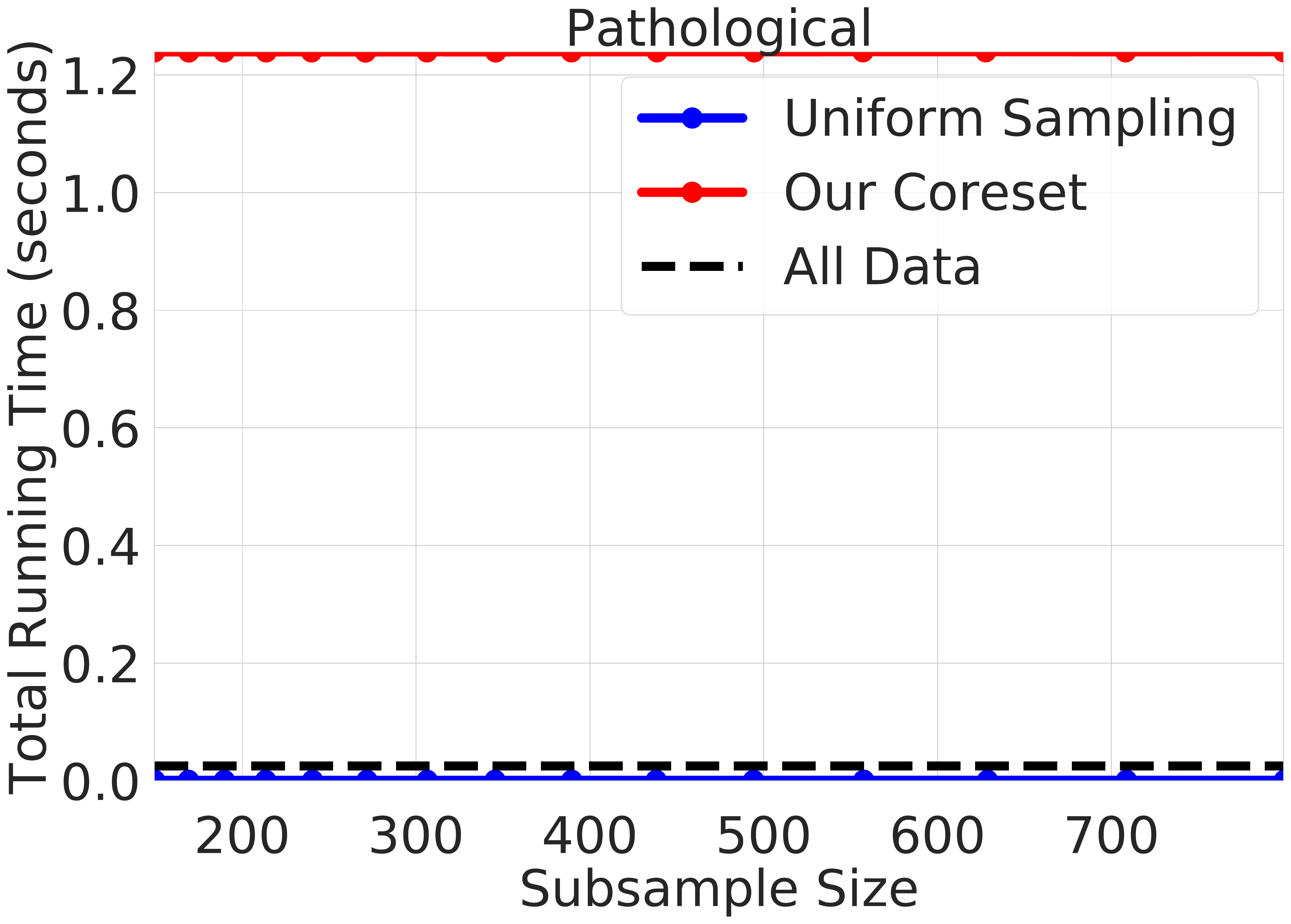}
  \end{minipage}%
  
\begin{minipage}[b]{0.32\textwidth}
  \centering
 \includegraphics[width=1\textwidth]{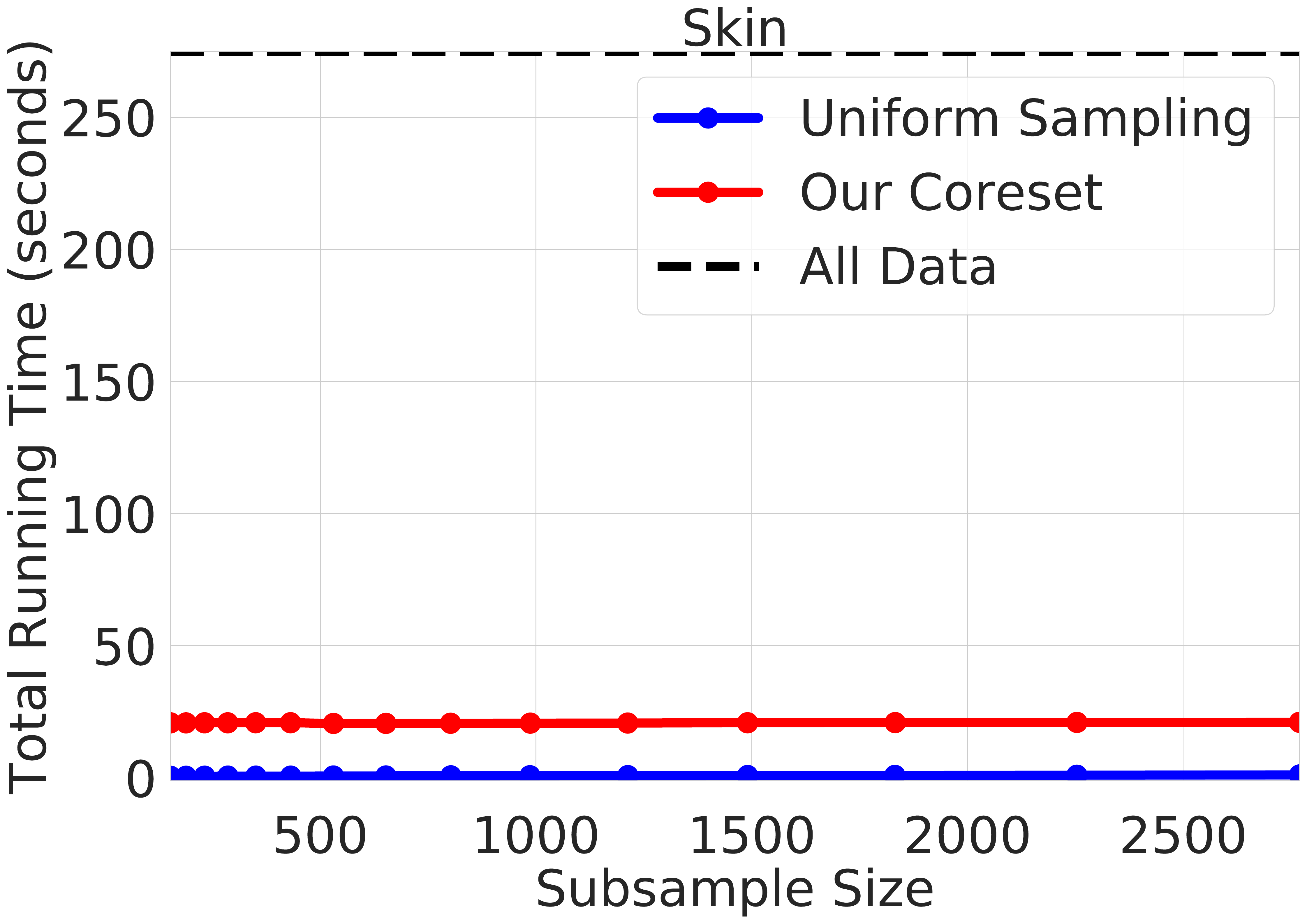}
  \end{minipage}%
  \begin{minipage}[b]{0.32\textwidth}
  \centering
 \includegraphics[width=1\textwidth]{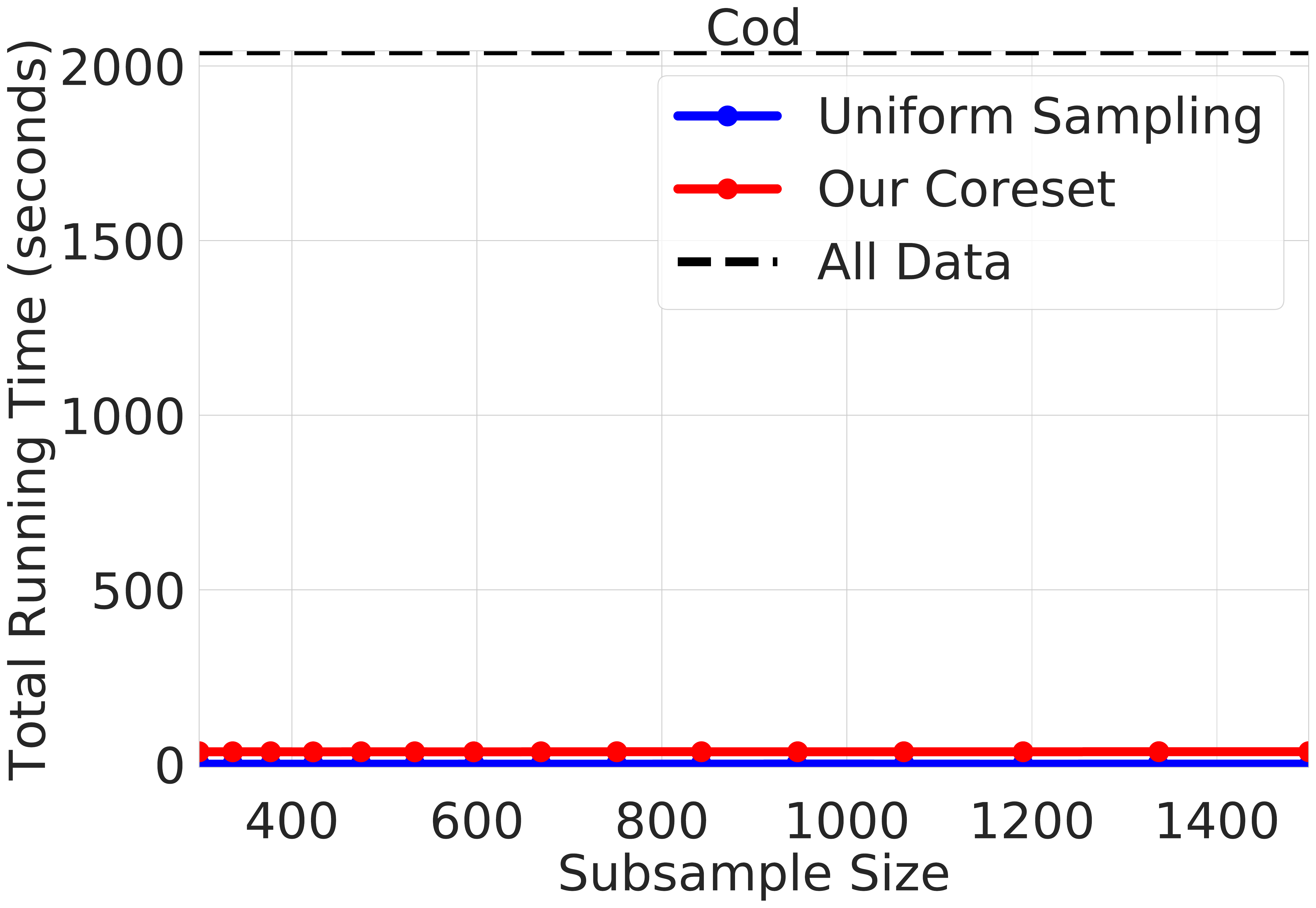} 
   \end{minipage}%
  \begin{minipage}[b]{0.32\textwidth}
  \centering
 \includegraphics[width=1\textwidth]{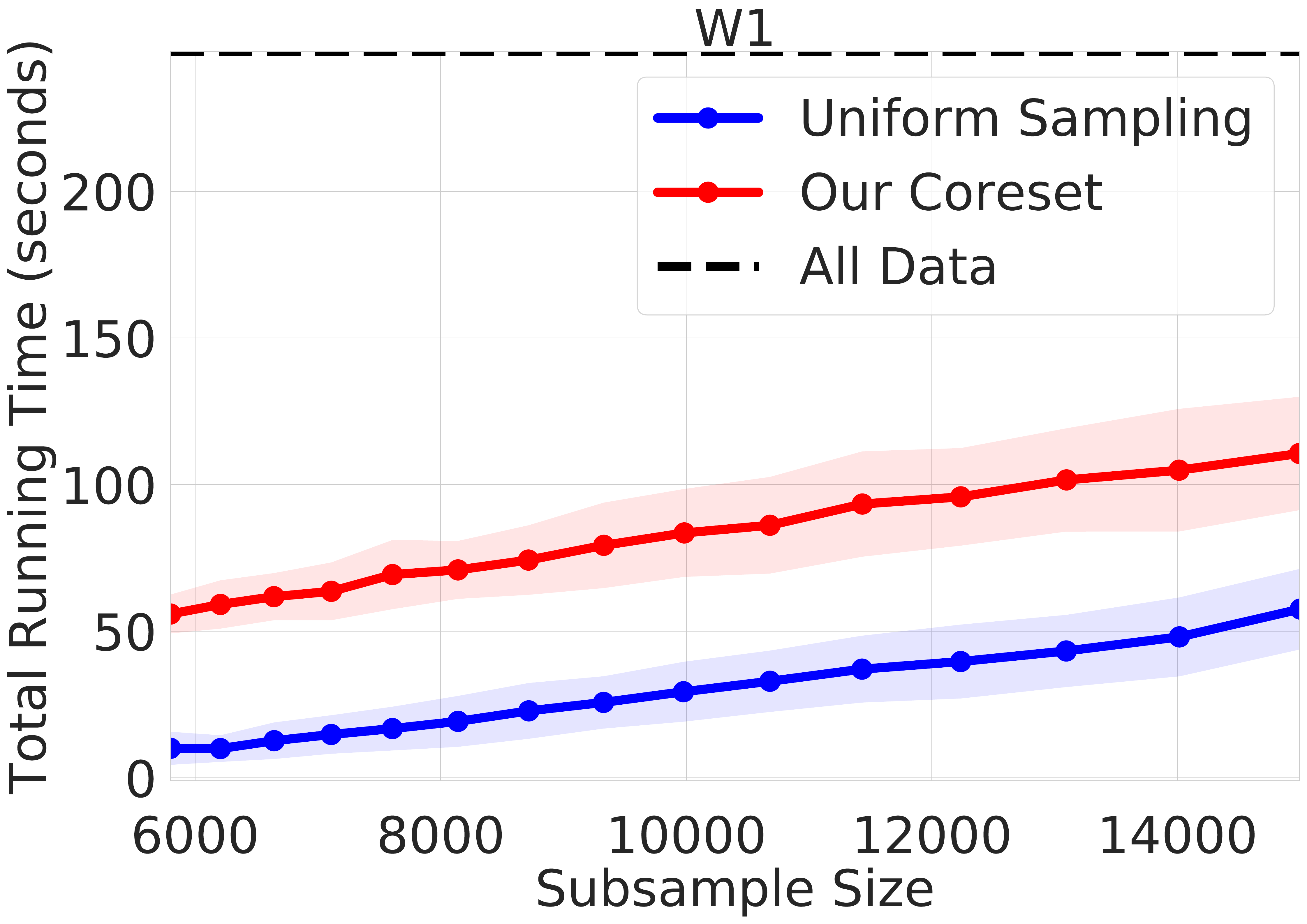}
  \end{minipage}%

    \caption{The \emph{total} computational cost of constructing a coreset and training the SVM model on the coreset, plotted as a function of the size of the coreset.}
	\label{fig:timing}
\end{figure*}

Figures~\ref{fig:RelativeError} and \ref{fig:timing} depict the results of our comparisons against uniform sampling in the offline setting. In Fig.~\ref{fig:RelativeError}, we see that the coresets generated by our algorithm are much more representative and compact than the ones constructed by uniform sampling: across all data sets and sample sizes, training on our coreset yields significantly better solutions to SVM problem when compared to those generated by training on a uniform sample. For certain data sets, such as HTRU, Pathological, and W1, this relative improvement over uniform sampling is at least an order of magnitude better, especially for small sample sizes. Fig.~\ref{fig:RelativeError} also shows that, as a consequence of a more informed sampling scheme, the variance of each model's performance trained on our coreset is much lower than that of uniform sampling for all data sets. 

Fig.~\ref{fig:timing} shows the total computational time required for constructing the sub-sample (i.e., coreset) $\SS$ and training the SVM on the subset $\SS$ to obtain $w_\SS^*$. We observe that our approach takes significantly less time than training on the original model when considering non-trivial data sets (i.e., $n \ge 18,000$), and underscores the efficiency of our method: we incur a negligible cost in the overall SVM training time due to a more involved coreset construction procedure, but benefit heavily in terms of the accuracy of the models generated (Fig.~\ref{fig:RelativeError}).

\begin{figure*}[!htb]
  \centering
  
  \begin{minipage}[b]{0.32\textwidth}
  \centering
 \includegraphics[width=1\textwidth]{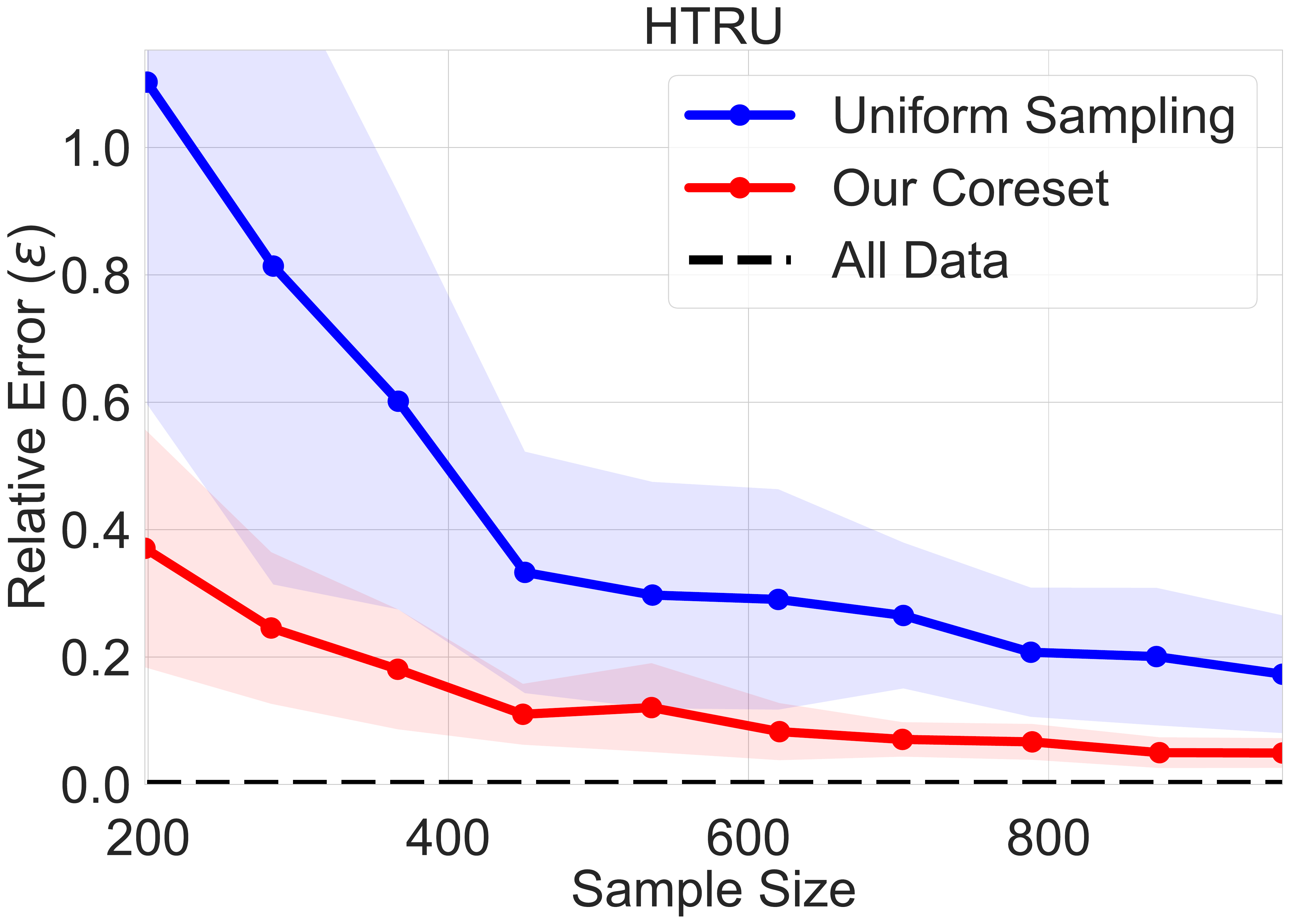}
  \end{minipage}%
  \begin{minipage}[b]{0.32\textwidth}
  \centering
 \includegraphics[width=1\textwidth]{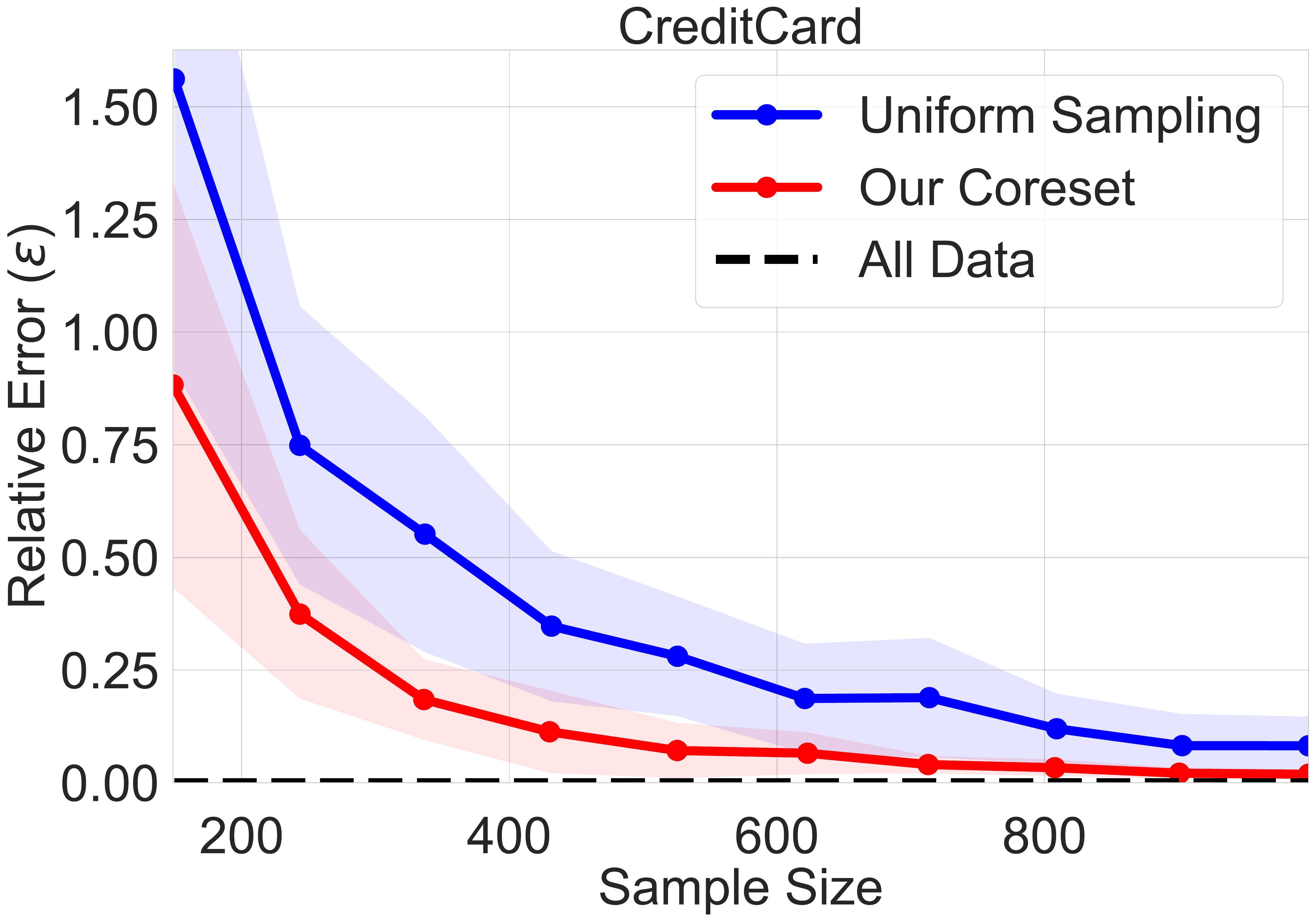}
  \end{minipage}%
  \begin{minipage}[b]{0.32\textwidth}
  \centering
 \includegraphics[width=1\textwidth]{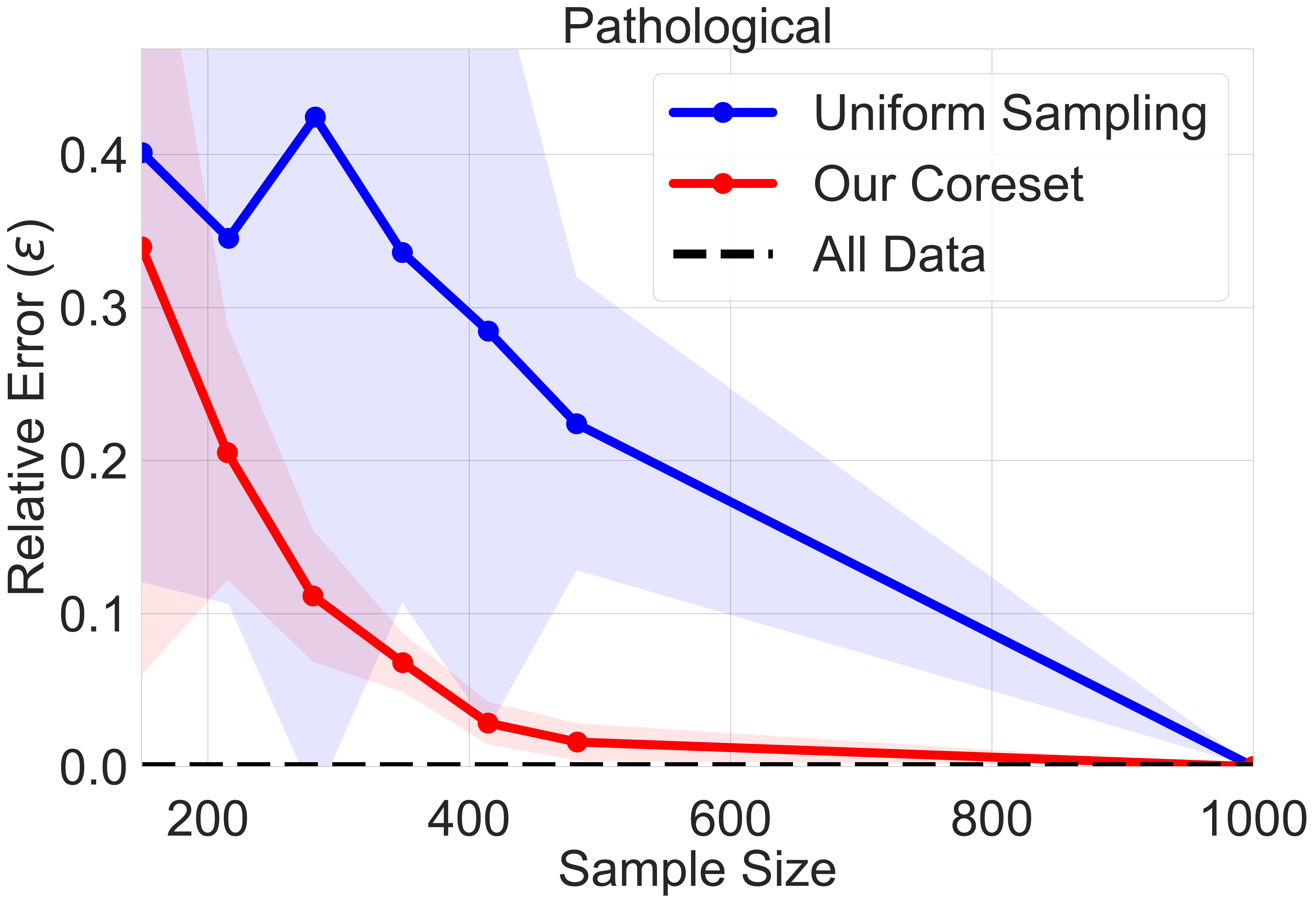}
  \end{minipage}%
  
\begin{minipage}[b]{0.32\textwidth}
  \centering
 \includegraphics[width=1\textwidth]{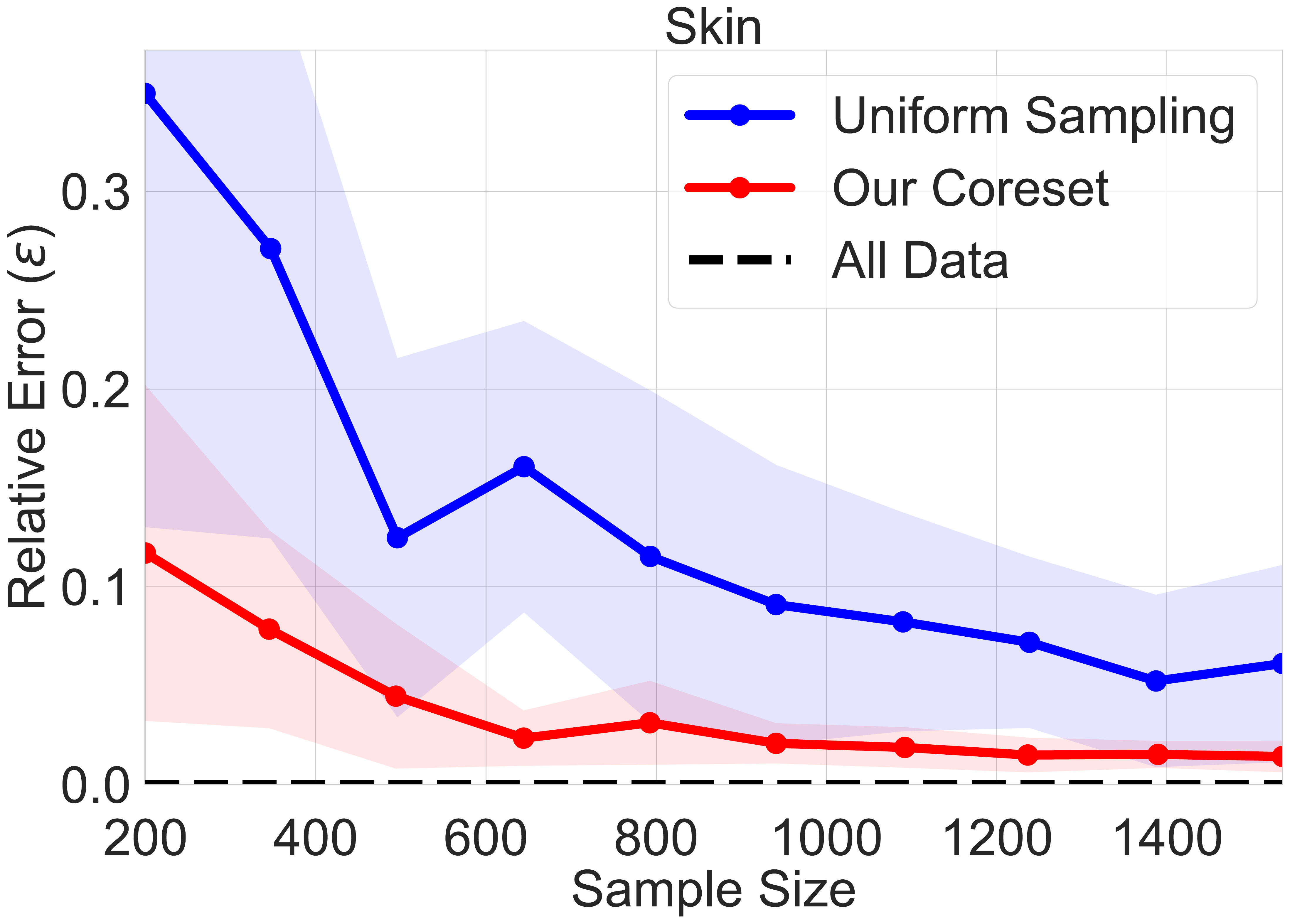}
  \end{minipage}%
\begin{minipage}[b]{0.32\textwidth}
  \centering
 \includegraphics[width=1\textwidth]{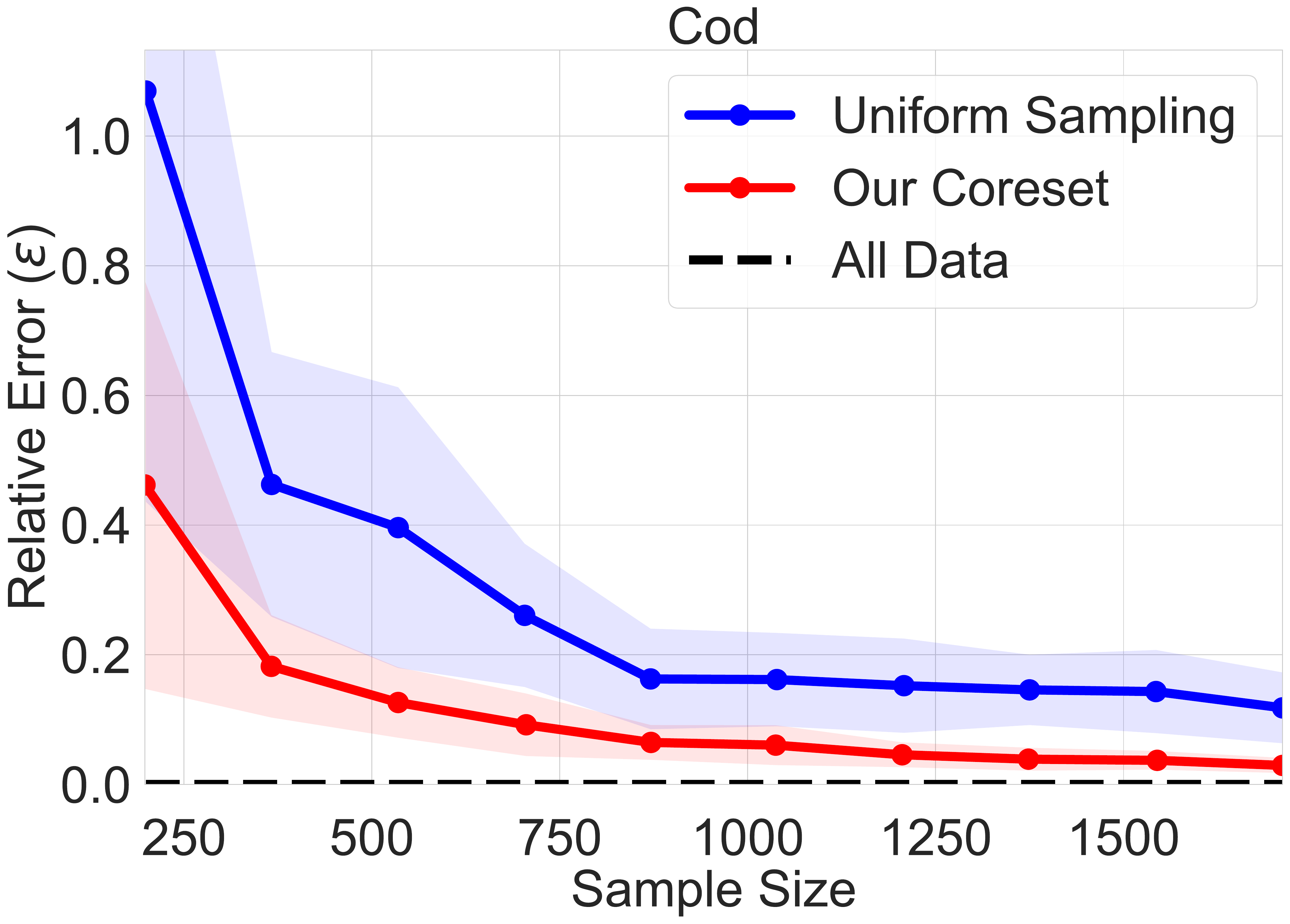}
  \end{minipage}%
  \begin{minipage}[b]{0.32\textwidth}
  \centering
 \includegraphics[width=1\textwidth]{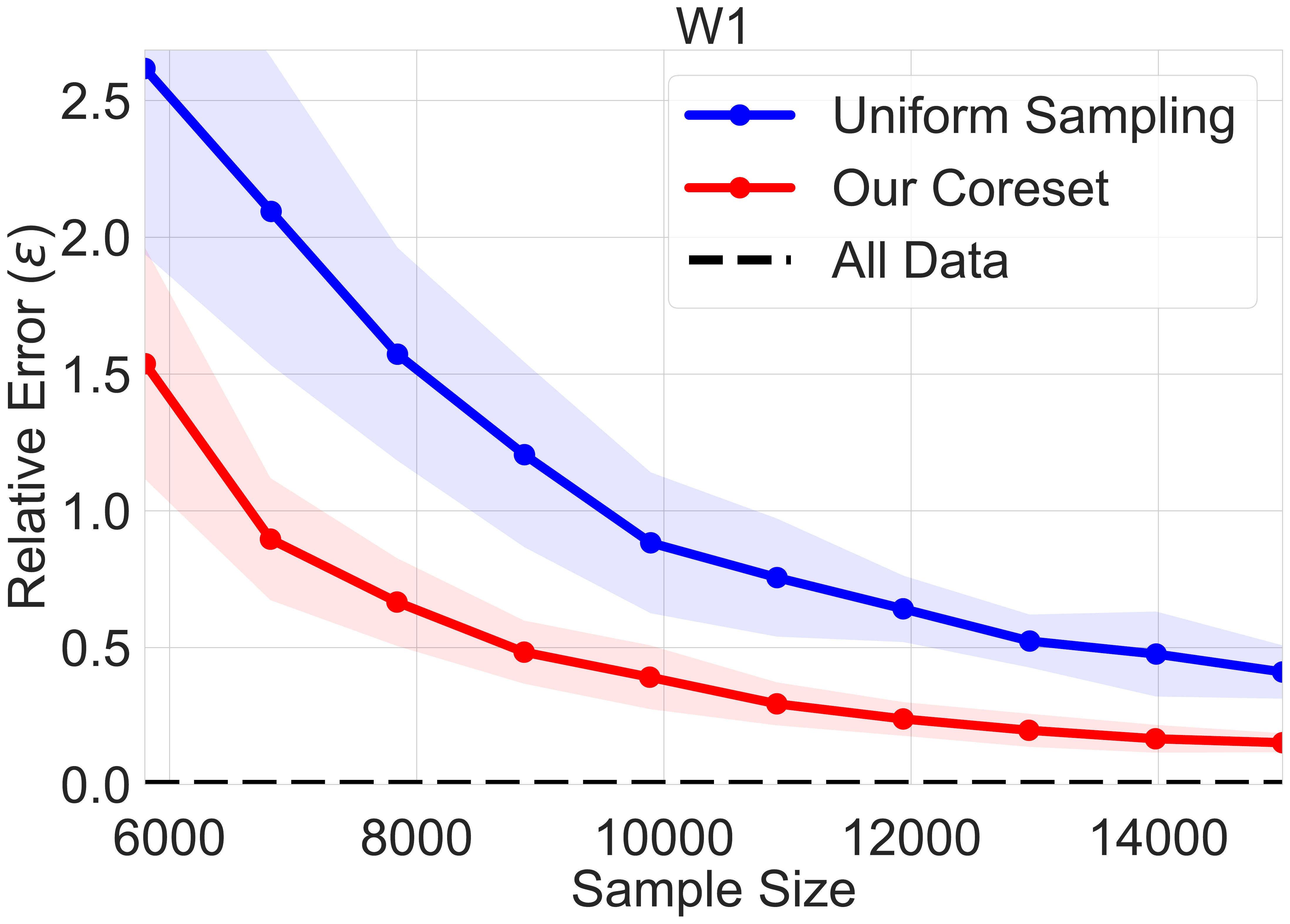} 
   \end{minipage}%

    \caption{The relative error of query evaluations with respect uniform and coreset subsamples for the $6$ data sets in the streaming setting. The figure shows that our method tends to fare even better in the streaming setting (cf. Fig.~\ref{fig:RelativeError}).}
	\label{fig:streaming}
\end{figure*}

Next, we evaluate our approach in the streaming setting, where data points arrive one-by-one and the entire data set cannot be kept in memory, for the same $6$ data sets. The results of the streaming setting are shown in Fig.~\ref{fig:streaming}. The corresponding figure for the total computational time is shown as Fig.~\ref{fig:streaming-timing} in Sec.~\ref{sec:appendix-Timing} of the Appendix.
Figs.~\ref{fig:streaming} and~\ref{fig:streaming-timing} (in the Appendix) portray a similar trend as the one we observed in our offline evaluations: our approach significantly outperforms uniform sampling for all of the evaluated data sets and sample sizes, with negligible computational overhead.

In sum, our empirical evaluations demonstrate the practical efficiency of our algorithm and reaffirm the favorable theoretical guarantees of our approach: the additional computational complexity of constructing the coreset is negligible relative to that of uniform sampling, and the entire preprocess-then-train pipeline is significantly more efficient than training on the original massive data set.

\section{Conclusion}
\label{sec:conclusion}
We presented an efficient coreset construction algorithm for generating compact representations of the input data points that are provably competitive with the original data set in training Support Vector Machine models. Unlike prior approaches, our method and its theoretical guarantees naturally extend to streaming settings and scenarios involving dynamic data sets, where points are continuously inserted and deleted. We established instance-dependent bounds on the number of samples required to obtain accurate approximations to the SVM problem as a function of input data complexity and established dataset dependent conditions for the existence of compact representations. Our experimental results on real-world data sets validate our theoretical results and demonstrate the practical efficacy of our approach in speeding up SVM training. We conjecture that our coreset construction can be extended to accelerate SVM training for other classes of kernels and can be applied to a variety of Big Data scenarios.


\bibliographystyle{splncs04}
\bibliography{main}

\appendix



\section{Proofs of the Analytical Results in Section~\ref{sec:analysis}}
\label{sec:appendix}
This section includes the full proofs of the technical results given in Sec.~\ref{sec:analysis}.

\subsection{Proof of Lemma~\ref{lem:sens-lower-bound}}
\label{sec:lem-sens-lower-bound-proof}

\senslowerbound*
\begin{proof}
Following \cite{yang2015weighted}, let $n = \binom{d}{d/2}$ and let $\PP = (P,u)$, where $P \subseteq \REAL^{d+1} \times \br{\pm 1}$ be set of $n$ labeled points, and $u : P \to 1$. For every $p=(x,y) \in P$, where $x \in \REAL^d \times \br{1}$ and $y \in \br{\pm 1}$, among the first $d$ entries of $x$, exactly $\frac{d}{2}$ entries are equivalent to
$$
y\sqrt{\frac{2}{d}},
$$
where the remaining $\frac{d}{2}$ entries among the first $d$ are set to $0$. 
Hence, for our proof to hold, we assume that $P$ contains all such combinations and at least one point of each label. For every $p=(x,y) \in P$, define the set of non-zero entries of $p$ as the set 
$$
B_p = \{i \in [d+1] : x_{i} \neq 0 \}.
$$

Put $p \in P$ and note that for bounding the sensitivity of point $p$, consider $w$ with entries defined as
$$
\forall{i \in [d+1]} \tab w_{i} = \begin{cases}
0 \tab \text{if} \tab i \in B_p, \\
\frac{1}{\sqrt{\frac{2}{d}}} \tab \text{otherwise}.
\end{cases}
$$

Note that $\norm{w}_2^2 = \frac{d}{2} \left( \frac{1}{\sqrt{\frac{2}{d}}} \right)^2 = \frac{d^2}{4}$. We also have that $h(p, w) = 1$ since $y \dotp{x}{w} = \sum_{i \in B_p} yx_{i} w_{i} = \frac{d}{2} 0 = 0$. To bound the sum of hinge losses contributed by other points $q \in P \setminus \br{p}$, note that $B_q \setminus B_p \neq \emptyset$. Then for every $q=(x^\prime,y^\prime)\neq p$,
\begin{align*}
y^\prime \dotp{x^\prime}{w} &= \sum_{i \in B_q \setminus B_p} y^\prime x^\prime_i w_i \ge \frac{1}{\sqrt{\frac{2}{d}}} \sqrt{\frac{2}{d}} = 1,
\end{align*}
which implies that $h(q, w) = 0$. Thus,
\[
\sum_{q \in P} h(q, w) = 1.
\]
Putting it all together, 
\begin{align*}
s(p) = \sup_{\substack{w^\prime \in \REAL^{d+1} \\ \F[w^\prime] \neq 0}} \frac{\f[w^\prime]}{\F[w^\prime]} \ge \frac{\frac{d^2}{8n} + \lambda \, h(p, w)}{\frac{\norm{w}_2^2}{2} + \lambda} = \frac{\frac{d^2}{8n} + \lambda}{\frac{d^2}{8} + \lambda}.
\end{align*}

Since the above holds for every $p \in P$, summing the above inequality over every $p \in P$, yields that
\begin{align*}
\sum_{p \in P} s(p) \ge \frac{\frac{d^2}{8} + n\lambda}{\frac{d^2}{8} + \lambda} \in \Omega \left(\frac{d^2 + n\lambda}{d^2 + \lambda}\right).
\end{align*}
\end{proof}



\subsection{Proof of Lemma~\ref{lem:sens-upper-bound}}
\label{sec:lem-sens-upper-bound-proof}
\sensupperbound*
\begin{proof}
Let $P_y \subseteq P$ denote the set of points with the same label as $p$ as in Line~\ref{alg:L4} of Algorithm~\ref{algorithm}. Consider an optimal clustering of the points in $P_y$ into $k$ clusters with centroids $\C^y = \{c_y^{(1)}, \ldots, c_y^{(k)}\} \subseteq \RD$ being their mean as in Line~\ref{alg:L5}, and let $\ap$ be as defined in Line~\ref{alg:alpha} for every $i \in [k]$ and $y \in \br{+,-}$. In addition, let $\PP \setminus \PP_y^{(i)}$ denote the weighted set $\left( P \setminus P_y^{(i)}, u \right)$ for every $i \in [k]$ and $y \in \br{+,-}$.

Put $p = (x,y) \in P$ and let $i\in [k]$ be the index of the cluster which $p$ belongs to, i.e., $p \in P_y^{(i)}$. 

We first observe that for any scalars $a, b \in \REAL$, $\max\{a - b, 0\} \leq \max\{a, 0\} + \max\{-b, 0\}$. This implies that, by definition of the hinge loss, we have for every $q, \hat{q}, w \in \RD$
$$
h(q, w) \leq h(\hat{q}, w) + \ramp{\dotp{q - \hat{q}}{w}},
$$
where $\ramp{x} = \max \{0, x\}$ as before. Hence, in the context of the definitions above
\begin{align}
\label{eq:hingeProp}
h(p, w) &= h(p - c(p) + c(p), w) \\
&\leq h(c(p), w) + \ramp{ \dotp{c(p) - y x}{w}} \\
&= h(c(p), w) + \ramp{\dotp{p_\Delta}{w}}.
\end{align}

Now let the total weight of the points in $P_y^{(i)}$ be denoted by $\U[P_y^{(i)}] = \sum\limits_{q \in P_y^{(i)}} u(q)$. Note that since $c_y^{(i)}$ is the centroid of $P_y^{(i)}$ (as described in Line~\ref{alg:L5} of Algorithm~\ref{algorithm}), we have
$
P_y^{(i)} = \frac{1}{\U[P_y^{(i)}]} \sum\limits_{q=(x_q,y_q) \in P_y^{(i)}} u(q) y_qx_q.
$
Observing that the hinge loss is convex, we invoke Jensen's inequality to obtain
$$
f_\lambda(c_y^{(i)}, w ) \leq \frac{1}{\U[P_y^{(i)}]} \sum_{q \in P_y^{(i)}} u(q) f(q, w) = \frac{\F[w] - F_\lambda(\PP \setminus \PP_y^{(i)}, w)}{\U[P_y^{(i)}]}.
$$

Applying the two inequalities established above to $s(p)/u(p)$ yields that
\begin{align}
\frac{s(p)}{u(p)} &= \sup_w \frac{\f}{\F[w]} \\
&\leq \sup_w \frac{f_\lambda(c_y^{(i)},w) + \lambda  \ramp{ \dotp{w}{p_\Delta}}}{\F[w]} \label{eq:bound_sens_1}\\
&\leq \sup_w \frac{\sum\limits_{q \in P_y^{(i)}} u(q) f_\lambda(q, w)}{ \U[P_y^{(i)}] \F[w]} + \frac{\lambda \ramp{ \dotp{w}{p_\Delta}}}{\F[w]} \\
&= \sup_w \frac{\F[w] - F_\lambda(\PP \setminus \PP_y^{(i)}, w)}{\U[P_y^{(i)}] \F[w]} + \frac{\lambda \ramp{ \dotp{w}{p_\Delta}}}{\F[w]} \\
&=\frac{1}{\U[P_y^{(i)}]} + \sup_w \frac{\lambda \ramp{ \dotp{w}{p_\Delta}} - F_\lambda(P \setminus P_y^{(i)}, w ) / \U[P_y^{(i)}]}{\F[w]} \label{eq:bound_sens_2} 
\end{align}

By definition of $F_\lambda(P \setminus P_y^{(i)}, w)$, we have
\begin{align*}
F_\lambda(P \setminus P_y^{(i)}, w) \ge \frac{\norm{w_{1:d}}_2^2 \,\U[P \setminus P_y^{(i)}]}{2 \U}.
\end{align*}

Continuing from above and dividing both sides by $\lambda$ yields
\begin{align*}
\frac{s(p)}{ \lambda u(p)} &\leq \frac{1}{\lambda \U[P_y^{(i)}]} + \sup_w \frac{\ramp{ \dotp{w}{p_\Delta}} - \frac{  \norm{w_{1:d}}^2 \, \U[P \setminus P_y^{(i)}]}{2 \lambda \U \U[P_y^{(i)}]}}{\F[w]}\\
&\leq \frac{1}{\lambda u(\C_p)} + \sup_w \frac{ \ramp{ \dotp{w}{p_\Delta}} - \ap \norm{w_{1:d}}_2^2}{\F[w]},
\end{align*}
where
\begin{equation}
\label{eq:alpha}
\ap = \alphaDefNew.
\end{equation}

Let
$$
g(w) = \frac{ \ramp{ \dotp{w}{p_\Delta}} - \ap \norm{w_{1:d}}_2^2}{\F[w]}
$$
be the expression on the right hand side of the sensitivity inequality above, and let $\hat w \in \argmax_{w} g(w)$. The rest of the proof will focus on bounding $g(\hat w)$, since an upper bound on the sensitivity of a point as a whole would follow directly from an upper bound on $g(\hat w)$.

Note that by definition of $p_\Delta$ and the embedding of $1$ to the $(d+1)^\text{th}$ entry of the original $d$-dimensional point (with respect to $p$),
\[
    \dotp{\hat w}{p_\Delta} = \dotp{\hat w_{1:d}}{(p_\Delta)_{1:d}},
\]
where the equality holds since the $(d+1)$th entry of $p_\Delta$ is zero. 

We know that $\dotp{\hat w}{p_\Delta} \ge \ap \norm{\hat w_{1:d}}_2^2 \ge 0$, since otherwise $g(\hat w) < 0$, which contradicts the fact that $\hat w$ is the maximizer of $g(w)$. This implies that for each entry $j \in [d]$ of the sub-gradient of $g(\cdot)$ evaluated at $\hat w$, denoted by $\nabla g(\hat w)$, is given by
\begin{align}
\label{eq:gradZero}
\nabla g(\hat{w})_j &= \frac{\left( (p_{\Delta})_j - 2 \ap \hat w_j \right) \F - \nabla \F_j \left( \dotp{w}{p_\Delta} - \ap \norm{w_{1:d}}_2^2\right) }{{\F}^2},
\end{align}
and that $\nabla g(\hat{w})_{d+1} = 0$ since the bias term does not appear in the numerator of $g(\cdot)$.

Letting $\gamma = \dotp{\hat w}{p_\Delta} - \ap \norm{\hat w_{1:d}}_2^2$ and setting each entry of the gradient $\nabla g(\hat w)$ to $0$, we solve for $p_\Delta$ to obtain 
$$
(p_\Delta)_{1:d} = \frac{\gamma \, \nabla \F_{1:d}}{\F} + 2 \ap \hat w_{1:d}.
$$

This implies that
\begin{align*}
    \dotp{\hat w}{p_\Delta} &= \frac{\gamma \, \dotp{\hat w}{\nabla \F}}{\F} + 2 \ap \norm{\hat w_{1:d}}_2^2
\end{align*}
Rearranging and using the definition of $\gamma$, we obtain
\begin{equation}
\label{eqn:dot}
\gamma = \frac{\gamma \, \dotp{\hat w}{\nabla \F}}{\F} + \ap \norm{\hat w_{1:d}}_2^2,
\end{equation}
where Lemma~\ref{lem:sens-upper-bound} holds by taking $\frac{9}{2}$ outside the max term.





By using the same equivalency for $p_\Delta$ from above, we also obtain that
\begin{align*}
    \norm{p_\Delta}_2^2 &=\dotp{p_\Delta}{p_\Delta} = \norm{\frac{\gamma \, \nabla \F_{1:d}}{\F} + 2 \ap \hat w}^2 \\
    &= \frac{\gamma^2}{\F^2} \norm{\nabla \F}_2^2 + 4 \left(\ap \right)^2 \norm{\hat w_{1:d}}_2^2 +4\ap  \frac{\gamma \, \dotp{\hat w}{\nabla \F}}{\F},
\end{align*}
but $\frac{\gamma \, \dotp{\hat w}{\nabla \F}}{\F} = \gamma - \ap \norm{\hat w_{1:d}}_2^2$, and so continuing from above, we have
\begin{align*}
    \norm{p_\Delta}_2^2 &= \frac{\gamma^2}{\F^2} \norm{\nabla \F}_2^2 + 4 \left(\ap \right)^2 \norm{\hat w_{1:d}}_2^2 +4\ap  (\gamma - \ap \norm{\hat w_{1:d}}_2^2) \\
    &= \frac{\gamma^2}{\F^2} \norm{\nabla \F_{1:d}}_2^2 + 4 \ap \gamma \\
    &= \gamma^2 \Tilde{x} + 4 \ap \gamma,
\end{align*}
where $\Tilde{x} = \frac{\norm{\nabla \F}_2^2}{\F^2}$. Solving for $\gamma$ from the above equation yields for $\tilde x > 0$
\begin{equation}
\label{eqn:golden}
\gamma = \frac{\sqrt{4 \left(\ap \right)^2 + \norm{p_\Delta}^2 \Tilde{x}} - 2 \ap}{\Tilde{x}}.
\end{equation}

Now we subdivide the rest of the proof into two cases. The first is the trivial case in which the sensitivity of the point is sufficiently small enough to be negligible, and the second case is the involved case in which the point has a high influence on the SVM cost function and its contribution cannot be captured by the optimal solution $w^*$ or something close to it.
\begin{description}[leftmargin=0pt, itemindent=20pt,
labelwidth=15pt, labelsep=5pt, listparindent=0.7cm,
align=left]

    \item[Case $g(\hat w) \leq 3 \ap$]
    the bound on the sensitivity follows trivially from the analysis above.

    \item[\label{case:A-norm} Case $g(\hat w) > 3\ap$]
    note that the assumption of this case implies that $w^*$ cannot be the maximizer of $g(\cdot)$, i.e., $\hat w \neq w^*$. This follows by the convexity of the SVM loss function which implies that the norm of the gradient evaluated at $w^*$ is 0. Thus by~\eqref{eqn:dot}:
    $$
    \gamma = \ap \w[w^*]^2.
    $$
    Since $\F[w^*] \ge \w[w^*]^2/2$, we obtain
    $$
    s(p) \leq \frac{\ap \w[w^*]^2}{\F[w^*]} \leq 2 \ap.
    $$
    Hence, we know that for this case we have $\grad[w^*] > 0$, $\F > \F[w^*] \ge 0$, and so we obtain $\x > 0$.
    
    This implies that we can use Eq.\eqref{eqn:golden} to upper bound the numerator $\gamma$ of the sensitivity. Note that $\gamma$ from \eqref{eqn:golden} is decreasing as a function of $\x$, and so it suffices to obtain a lower bound on $\x$. To do so, lets focus on Eq.\eqref{eqn:dot} and let divide both sides of it by $\gamma$, to obtain that
    \[
    1  = \frac{\dotp{\hat w}{\nabla \F}}{\F} + \frac{\ap}{\gamma} \norm{w_{1:d}}_2^2.
    \]
    
    By rearranging the above equality, we have that
    \begin{equation}
    \label{eq:gradFOverF}
    \frac{\dotp{\hat w}{\nabla \F}}{\F} = 1 - \frac{\ap\norm{w_{1:d}}_2^2}{\gamma}.
    \end{equation}

    Recall that since the last entry of $p_\Delta$ is $0$ then it follows from Eq.\eqref{eq:gradZero} that $\nabla \F_{d+1}$ is also zero, which implies that
    \begin{equation}
    \label{eq:wDotGrad}
    \begin{split}
    \dotp{\hat w}{\nabla \F} &= \dotp{\hat w_{1:d}}{\nabla \F_{1:d}} \\
    &\leq \norm{\hat w_{1:d}}_2 \norm{\nabla \F_{1:d}}_2 \\
    &= \norm{\hat w_{1:d}}_2 \norm{\nabla \F}_2
    \end{split}
    \end{equation}
    where the inequality is by Cauchy-Schwarz.
    
    Combining Eq.\eqref{eq:gradFOverF} with Eq.~\eqref{eq:wDotGrad} yields
    \begin{align*}
    \frac{\norm{\hat w_{1:d}}_2 \norm{\nabla \F}_2}{\F} &\geq 1 - \frac{\ap \norm{w_{1:d}}_2^2}{\gamma} \\ 
    &\geq 1 - \frac{\ap \norm{\hat w_{1:d}}_2^2}{3\ap \F} \\
    &\geq 1 - \frac{\ap 2 \F}{3 \ap \F} \\
    &= \frac{1}{3},
    \end{align*}
    where the second inequality holds by the assumption of the case, the third inequality follows from the fact that $\norm{\hat w_{1:d}}_2^2 \leq 2 \F$.
    
    This implies that 
    \begin{align*}
    \frac{\norm{\nabla \F}_2}{\F} \geq \frac{1}{3 \norm{w_{1:d}}_2} \geq \frac{\sqrt{2}}{3 \sqrt{\F}}. 
    \end{align*}
    
    Hence by definition of $\Tilde{x}$, we have that 
    \begin{equation}
    \label{eq:gradOverFLowerBound}
    \Tilde{x} \geq \frac{2}{9 \F}
    \end{equation}
    
    Plugging Eq.\eqref{eq:gradOverFLowerBound} into Eq.\eqref{eqn:golden}, we obtain that 
    \[
    \frac{\gamma}{\F} \leq \frac{9}{2} \left( \sqrt{4\left( \ap\right)^2 + \frac{2\norm{p_\Delta}_2^2}{9\F}} -2\ap\right).
    \]
    
    Recall that
    \[
    \F \geq \F[w^*] \geq \F[\Tilde{w}] - \xi,
    \]
    which implies that
    \begin{equation}
    \label{eq:bound_sens1}
    \frac{\gamma}{\F} \leq \frac{9}{2} \left( \sqrt{4\left( \ap\right)^2 + \frac{2\norm{p_\Delta}_2^2}{9\OPT_\xi}} -2\ap\right),
    \end{equation}
    where $\OPT_\xi = \F[\Tilde{w}] - \xi$.
\end{description}

Combining both cases, yields that
\begin{equation}
\label{eq:final_sens_bound}
s(p) \leq \frac{u(p)}{\U[P_y^{(i)}]} + u(p)\lambda \, \max\br{2 \ap, \frac{9}{2} \left( \sqrt{4\left( \ap\right)^2 + \frac{2\norm{p_\Delta}_2^2}{9\OPT_\xi}} -2\ap\right)},
\end{equation}
where Lemma~\ref{lem:sens-upper-bound} holds by rearranging Eq.~\ref{eq:final_sens_bound}.
\end{proof}

\subsection{Proof of Lemma~\ref{lem:sum-sens-upper-bound}}
\label{sec:lem-sum-upper-bound-proof}
\setcounter{theorem}{\getrefnumber{lem:sum-sens-upper-bound}-1}
\sumsensupperbound*

\begin{proof}
We first observe that that
\[
\sum\limits_{p \in P} s(p) = \sum\limits_{i \in [k]} \left( \sum\limits_{p \in P_+^{(i)}} s(p) + \sum\limits_{p \in P_{-}^{(i)}} s(p) \right).
\]

Thus we will focus on the summing the sensitivity of the all the points whose label is positive. We note that 
\begin{equation}
\label{eq:boundSumWeights}
\sum\limits_{i \in [k]} \sum\limits_{p \in P_{+}^{(i)}} \frac{u(p)}{\U[P_+^{(i)}]} = \sum\limits_{i=1}^k 1 = k.
\end{equation}

In addition, we observe that $\max\br{a,b} \leq a + b$ for every $a,b \geq 0$, which implies that for every $i \in [k]$ and $p \in P_{+}^{(i)}$,
\begin{equation}
\label{eq:max_term_splitted}
\begin{split}
&\max\br{2 \ap, \frac{9}{2} \left( \sqrt{4\left( \ap\right)^2 + \frac{2\norm{p_\Delta}_2^2}{9\OPT_\xi}} -2\ap\right)} \\
&\leq 2\ap + \frac{9}{2} \left( \sqrt{4\left( \ap\right)^2 + \frac{2\norm{p_\Delta}_2^2}{9\OPT_\xi}} -2\ap\right).
\end{split}
\end{equation}

Since $\sqrt{a + b} \leq \sqrt{a} + \sqrt{b}$ for every $a,b \geq 0$, we have for every $i \in [k]$ and $p \in P_{-}^{(i)}$,
\begin{equation}
\label{eq:boundSqrtTerm}
\begin{split}
&\sqrt{4\left( \ap\right)^2 + \frac{2\norm{p_\Delta}_2^2}{9\OPT_\xi}} -2\ap \\
&\leq 2\ap + \frac{\sqrt{2} \norm{p_\Delta}_2}{3 \sqrt{\OPT_\xi}} - 2\ap \\
&= \frac{\sqrt{2} \norm{p_\Delta}_2}{3 \sqrt{\OPT_\xi}}.
\end{split}
\end{equation}

Hence by combining Eq.\eqref{eq:boundSumWeights}, Eq.\eqref{eq:max_term_splitted}, and Eq.\eqref{eq:boundSqrtTerm}, we yield that
\begin{equation*}
\begin{split}
\sum\limits_{i \in [k]} \sum\limits_{p \in P_{+}^{(i)}} s(p) &\leq k + \sum\limits_{i \in [k]} \sum\limits_{p \in P_{+}^{(i)}} 2\lambda \ap[+] + \frac{9}{2}\lambda u(p) \frac{\sqrt{2}\norm{p_\Delta}_2}{3 \sqrt{\OPT_\xi}} \\
&= k + \sum\limits_{i \in [k]} \sum\limits_{p \in P_{+}^{(i)}} 2\lambda \ap[+] + \sum\limits_{i \in [k]} \lambda \frac{3 \text{Var}_+^{(i)}}{\sqrt{2 \OPT_\xi}} \\
&\leq 2k + \sum\limits_{i \in [k]} \lambda \frac{3 \text{Var}_+^{(i)}}{\sqrt{2 \OPT_\xi}},
\end{split}
\end{equation*}
where the inequality follows from definition of $\ap$ for every $i \in [k]$ and $y \in \br{+,-}$ as defined in Eq.\eqref{eq:alpha}.

Since all of the previous arguments hold similarly for $P_{-}$, we obtain that 
\[
\sum\limits_{p \in P} s(p) \leq 4k + \sum\limits_{i \in [k]} \frac{3 \text{Var}_+^{(i)}}{\sqrt{2 \OPT_\xi}} + \frac{3 \text{Var}_-^{(i)}}{\sqrt{2 \OPT_\xi}}.
\]
\end{proof}

\subsection{Proof of Theorem~\ref{thm:epsilon-coreset}}
\label{sec:thm-epsilon-coreset-proof}

\setcounter{theorem}{\getrefnumber{thm:epsilon-coreset}-1}
\epsiloncoreset*
\begin{proof}
By Lemma~\ref{lem:sens-upper-bound} and Theorem 5.5 of~\cite{braverman2016new} we have that the coreset constructed by our algorithm is an $\epsilon$-coreset with probability at least $1 - \delta$ if
$$
m \ge \SamplesNeeded,
$$
where we used the fact that the VC dimension of a SVMs in the case of a linear kernel is bounded $\text{dim}(\mathcal{F}) \leq d + 1 = \BigO(d)$~\cite{vapnik1998statistical}, and $c$ is a sufficiently large constant which can be determined using similar techniques to that of~\cite{li2001improved}. Moreover, note that the computation time of our algorithm is dominated by going over the whole weighted set $\PP$ which takes $\BigO\left( n\right)$ and attaining an $\xi$-approximation to the SVM problem at Line~\ref{alg:L1} followed by applying $k$-means clustering as shown in Algorithm~\ref{algorithm} which takes $\BigO \left( T\right)$ time. This implies that the overall time is $\BigO\left( nd + T \right)$.
\end{proof}

\begin{corollary}
\label{cor:error-bound}
Let $\PP$ be a weighted set, $\eps \in \left(0, \frac{1}{2} \right)$ and let $\SS$ be an $\eps$-coreset with respect to $\PP$. Let $w^*_{\SS} = \argmin_{w \in \REAL^{d+1}} F_\lambda\left(\SS, w \right)$ and let $w^*_{\PP}$  be defined similarly with respect to $\PP$.
\[
\F[w^*_{\PP}] \leq \F[w^*_{\SS}] \leq (1 + 4 \eps) \F[w^*_{\PP}]
\]
\end{corollary}

\begin{proof}
By Theorem~\ref{thm:epsilon-coreset}, $\SS=(S,v)$ is an $\epsilon$-coreset for $(\PP, u)$ with probability at least $1 - \delta$, which implies that
\begin{align*}
F_\lambda(\PP, w^*_{\SS}) &\leq \frac{F_\lambda(\SS, w^*_{\SS})}{1 - \epsilon} \leq \frac{F_\lambda(\SS, w^*_{\PP})}{1 - \epsilon} \\
&\leq \frac{1 + \epsilon}{1 - \epsilon} F_\lambda(\PP, w^*_{\PP}) \leq (1 + 4 \epsilon) F_\lambda(\PP, w^*_{\PP}),
\end{align*}
where the first and third inequalities follow from $(S,v)$ being an $\eps$-coreset (see Definition~\ref{def:epsCoreset}), the second inequality holds by definition of $w^*_{\SS}$, and the last inequality follows from the assumption that $\epsilon \in \left( 0, \frac{1}{2}\right)$.
\end{proof}

\paragraph{Sufficient Conditions} \label{par:sufficient}
Theorem~\ref{thm:epsilon-coreset} immediately implies that, for reasonable $\epsilon$ and $\delta$, coresets of poly-logarithmic (in $n$) size can be obtained if $d =\Bigo(\text{polylog}(n))$, which is usually the case in our target Big Data applications, and if $$\sum\limits_{i=1}^k \frac{3\lambda \text{Var}_+^{(i)}}{\sqrt{2\OPT_\xi}} + \frac{3\lambda \text{Var}_-^{(i)}}{\sqrt{2\OPT_\xi}} \allowbreak = \Bigo(\text{polylog}(n)).$$
For example, a value of $\lambda \leq \frac{\log n}{n}$ for the regularization parameter $\lambda$ satisfies the sufficient condition for all data sets with points normalized such that they are contained within the unit ball. Note that the total sensitivity, which dictates how many samples are necessary to obtain an $\epsilon$-coreset with probability at least $1 - \delta$ and in a sense measures the difficulty of the problem, increases monotonically with the sum of distances of the points from their label-specific means.

\section{Extension to Streaming Settings}
\label{sec:streaming-extension}
As a corollary to our main method, Alg.~\ref{algorithmStreams} extends the capabilities of any SVM solver, exact or approximate, to the streaming setting, where data points arrive one-by-one. Alg.~\ref{algorithmStreams} is inspired by~\cite{feldman2011unified,langberg2010universal} and constructs a binary tree, termed the \emph{merge-and-reduce tree}, starting from the leaves which represent chunks of the data stream points. For each stream of $l$ points, we construct an $\eps$-coreset using Algorithm~\ref{algorithm}, and then we add each resulted tuple to $B_1$, a bucket which is responsible for storing each of the parent nodes of the leaves (Lines 1-6).

Note that for every $i > 1$, the bucket $B_i$ will hold every node which is a root of a subtree of height $i$. 
Then, for every two successive items in each bucket $B_i$, for every $i \geq 1$, a parent node is generated by computing a coreset on the union of the coresets, which is then, added to the bucket $B_{i+1}$ (Lines 7-11). 
This process is done till all the buckets are emptied other than $B_h$, that will contain only one tuple $(S,v)$ which is set to be the root of the merge-and-reduce tree.

In sum, we obtain a binary tree of height $h = \Theta(\log (n))$ for a stream of $n$ data points. Thus, at Lines~\ref{alg2:L3.5} and~\ref{alg2:L9}, we have used error parameter $\epsilon' = \epsilon/(2\log(n))$ and failure parameter $\delta' = \delta / (2\log(n))$ in order to obtain $\eps$-coreset with probability at least $1 - \delta$.

\begin{algorithm}[!htb]
\SetKwInOut{Input}{Input}
\SetKwInOut{Output}{Output}
\Input{An input stream $P$ in $\REAL^{d+1} \times \{-1,1\}$ of $n$ points, a leaf size $\ell > 0$, a weight function $u : P \to \REAL_{\geq 0}$, a regularization parameter $\lambda \in [0,1]$, a positive integer $k$, and an approximation factor $\xi > 0$.}
\Output{A weighted set $(\SS,v)$}
\caption{$\streamCore(P,u,\ell,\lambda,\xi,k)$\label{two}}
\label{algorithmStreams}
\vskip -0.02in
$B_i \gets \emptyset$ for every $1 \leq i \leq \infty$; \label{alg2:L1} \\
$h \gets 1$; \label{alg2:L2} \\
\For{\label{alg2:L3} each set $Q$ of consecutive $2\ell$ points from $P$}{ 
    $(T,v) \gets \coreset(Q,u, \lambda, \xi, k, \ell)$; \, \,  $j \gets 1$; \label{alg2:L3.5}\\
    $B_j \gets B_j \cup (T,v)$; \label{alg2:L4}\\
    \For{\label{alg2:L5} each $j \leq h$}{
    	\While{$\abs{B_j} \geq 2$}{
        	$(T_1,u_1) , (T_2,u_2) \gets $ top two items in $B_j$; \\
        	Set $\Tilde{u} : T_1 \cup T_2 \to [0,\infty)$ such that for every $p \in T_1 \cup T_2$, 
        	$\Tilde{u}(p) = \begin{cases} u_1(p) & p \in T_1,\\
        	                              u_2(p) & \text{otherwise}\end{cases}$;\\
        	$(T,v) \gets \coreset(T_1 \cup T_2, \Tilde{u}, \lambda, \xi,  k, \ell)$ \label{alg2:L9}; \\
            $B_{j+1} \gets B_{j+1} \cup (T,v)$;\label{alg2:L6} \\ 
            $h \gets \max\{h, j+1\}$;\label{alg2:L7}\\
        }
    }
}
Set $(S,v)$ to be the only item in $B_h$ \\
\Return $(S,v)$ \label{alg2:L8}\\
\end{algorithm}

\paragraph{Coreset construction of size poly-logarithmic in $n$.}


In case of the total sensitivity being sub-linear in $n$ where $n$ denotes the number of points in $P$, which is obtained by Lemma~\ref{lem:sens-upper-bound}, we provide the following theorem which constructs a $(1+\eps)$-coreset of size poly-logarithmic in $n$.

\begin{restatable}{lemma}{mergeandreduce}
\label{thm:polylog-coreset-merge-reduce}
Let $\eps \in \left[ \frac{1}{\log{n}},\frac{1}{2} \right]$, $\delta \in \left[\frac{1}{\log{n}},1 \right)$, $\lambda \in (0,1]$, a weighted set $(P,u)$, $\xi \in \left[ 0, \F[w^*] \right]$ where $w^* \in \argmin_{w \in \REAL^{d+1}} \F[w]$. Let
$t$ denote the total sensitivity from Lemma~\ref{lem:sens-upper-bound} and suppose that there exists $\beta \in (0.1, 0.8)$ such that $t \in \Theta(n^\beta)$. Let $\ell \geq \max\br{2^{\frac{\beta}{1-\beta}}, \SamplesNeededStream}$ and let $(S,v)$ be the output of a call to $\streamCore(P, u, \ell, \lambda,\xi,)$. Then $(S,v)$ is an $\eps$-coreset of size 
\[
\abs{S} \in  \left( \log{n} \right)^{\Bigo \left( 1 \right)}.
\]
\end{restatable}
\begin{proof}
First we note that using Theorem~\ref{thm:epsilon-coreset} on each node in the merge-and-reduce tree, would attain that the root of the tree, i.e., $(S,v)$ attains that for every $w$
\[
(1-\eps)^{\log{n}} \F[w] \leq F_\lambda\left( (S,v), w \right) \leq (1+\eps)^{\log{n}} \F[w],
\]
with probability at least $(1-\delta)^{\log{n}}$.

We observe by the properties of the natural number $e$,
\begin{equation*}
(1+\eps)^{\log{n}} = \left(1 + \frac{\eps \log{n}}{\log{n}} \right)^{\log{n}} \leq e^{\eps\log{n}},
\end{equation*}
which when replacing $\eps$ with $\eps^\prime = \frac{\eps}{2 \log{n}}$ in the above inequality as done at Lines~\ref{alg2:L3.5} and~\ref{alg2:L9} of Algorithm~\ref{algorithmStreams}, we obtain that
\begin{align}
    (1+\eps^\prime)^{\log{n}} \leq e^{\frac{\eps}{2}} \leq 1 + \eps,
\end{align}
where the inequality holds since $\eps \in [\frac{1}{\log{n}}, \frac{1}{2}]$.

As for the lower bound, observe that
\[
(1-\eps)^{\log{n}} \geq 1 - \eps\log{n},
\]
where the inequality holds since $\eps \in [\frac{1}{\log{n}}, \frac{1}{2}]$.

Hence,
\[
(1-\eps^\prime)^{\log{n}} \geq 1 - \eps^\prime\log{n} = 1 - \frac{\eps}{2} \geq 1 - \eps.
\]

Similar arguments holds also for the failure probability $\delta$. What is left for us to do is setting the leaf size which will attain us an $\eps$-coreset of size poly-logarithmic in $n$ (the number of points in $P$).

Let $\ell \in (0,\infty)$ be the size of a leaf in the merge-and-reduce tree. We observe that a coreset of size poly-logarithmic in $n$, can be achieved by solving the inequality
\begin{align*}
\frac{2\ell}{2} \geq (2\ell)^\beta,
\end{align*}
which is invoked when ascending from any two leafs and their parent node at the merge-and-reduce tree.

Rearranging the inequality, we yield that
\begin{align*}
\ell^{1-\beta} \geq 2^\beta.
\end{align*}

Since $\ell \in (0,\infty)$, any $\ell \geq \sqrt[1-\beta]{2^\beta}$ would be sufficient for the inequality to hold. What is left for us to do, is to show that when ascending through the merge-and-reduce tree from the leaves towards the root, each parent node can't be more than half of the merge of it's children (recall that the merge-and-reduce tree is built in a binary tree fashion, as depicted at Algorithm~\ref{algorithmStreams}).

Thus, we need to show that,
\begin{align*}
2^{\sum\limits_{j=1}^{i} \beta^j} \cdot \ell^{\beta^i} \leq \frac{2^{\sum\limits_{k=0}^{i-1} \beta^k} \cdot \ell^{\beta^{i-1}}}{2} = 2^{\sum\limits_{k=1}^{i-1} \beta^k} \cdot \ell^{\beta^{i-1}},
\end{align*}
holds, for any $i \in \left[\ceil{\log{n}} \right]$ where $\log{n}$ is the height of the tree. Note that the left most term is the parent node's size and the right most term represents half the size of both parent's children nodes.

In addition, for $i = 1$, the inequality above represents each node which is a parent of leaves. Thus, we observe that for every $i \geq 1$, the inequality represents ascending from node which is a root of a sub-tree of height $i-1$ to it's parent in the merge-and-reduce tree. 

By simplifying the inequality, we obtain the same inequality which only addressed the leaves. Hence, by using any $\ell \geq 2^{\frac{\beta}{1-\beta}}$ as a leaf size in the merge and reduce tree, we obtain an $\eps$-coreset of size poly-logarithmic in $n$. 
\end{proof}

\section{The logic behind applying $k$-means clustering}
\label{sec:logic-kmeans}
First put in mind the bound from Lemma~\ref{lem:sens-upper-bound}, and note that it was achieved by using $k$-means++ clustering as depicted at Algorithm~\ref{algorithm}. Following our analysis from Sec.~\ref{sec:lem-sens-upper-bound-proof}, we observe that we can simply use any $k$-partitioning of the dataset, instead of applying $k$-means clustering. Moreover, we can also simply choose $k=1$ which translates to simply taking the mean of the labeled points. Despite all of above, we did choose to use a clustering algorithm as well as having larger values for $k$ and such decisions were inspired by the following observations:
\begin{enumerate}[label=(\roman*)]
    \item $k$-means clustering aims to optimize the sum of squared distances between the points and their respected center, which on some level, helps in lowering the distance between points and their respected centers. Such observation leads to having tighter bound for the sensitivity of each point, consequently leading to lower coreset sizes; See Thm.~\ref{thm:epsilon-coreset}.
    \item Having larger $k$ also helps lowering the distance between a point and its respected center.
    \item $k$-means clustering acts as a trade-off mechanism between
    \begin{itemize}[label=$\bullet$]
        \item the \emph{raw contribution} of each point, which is translated into the weight of the point divided by the sum of the weights with respect to the cluster that each point is assigned to, 
        \item and the \emph{actual contribution} of each point which is translated to the distance between each point and its respected center. 
    \end{itemize}
\end{enumerate}

\begin{figure*}[!htb]
  \centering
  \begin{minipage}[b]{0.45\textwidth}
  \centering
  \subcaptionbox{At first, an optimal solution for the SVM problem is found.}[1\textwidth]{\includegraphics[width=1\textwidth]{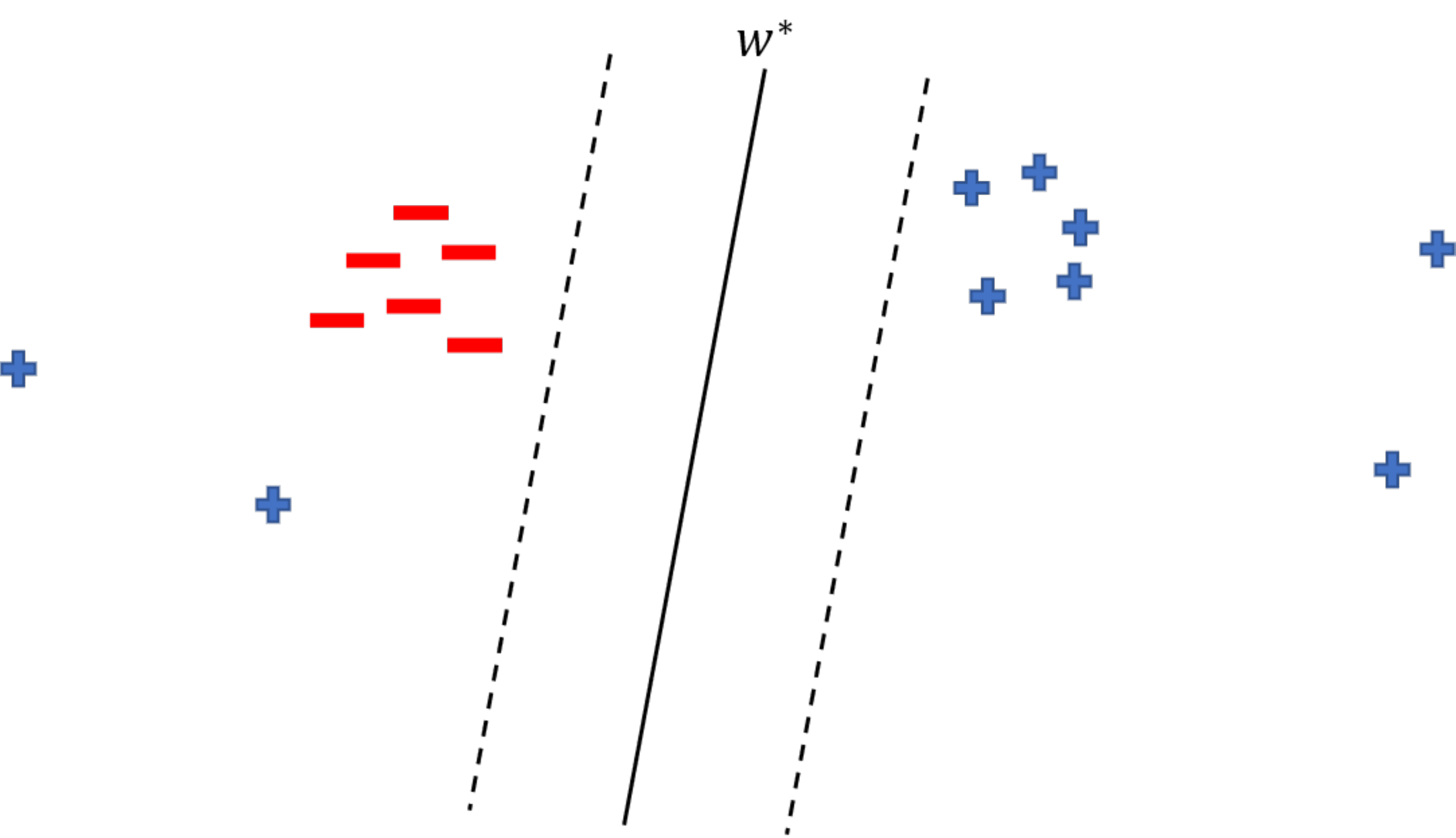}}
  \end{minipage}\qquad
  \begin{minipage}[b]{0.45\textwidth}
  \centering
 \subcaptionbox{We focus on the the positive labeled points (blue points) and we find a $k$-means clustering using $k$-means++, where we set $k=3$. The yellow points are the centers found by the $k$-means++ algorithm}[1\textwidth]{\includegraphics[width=1\textwidth]{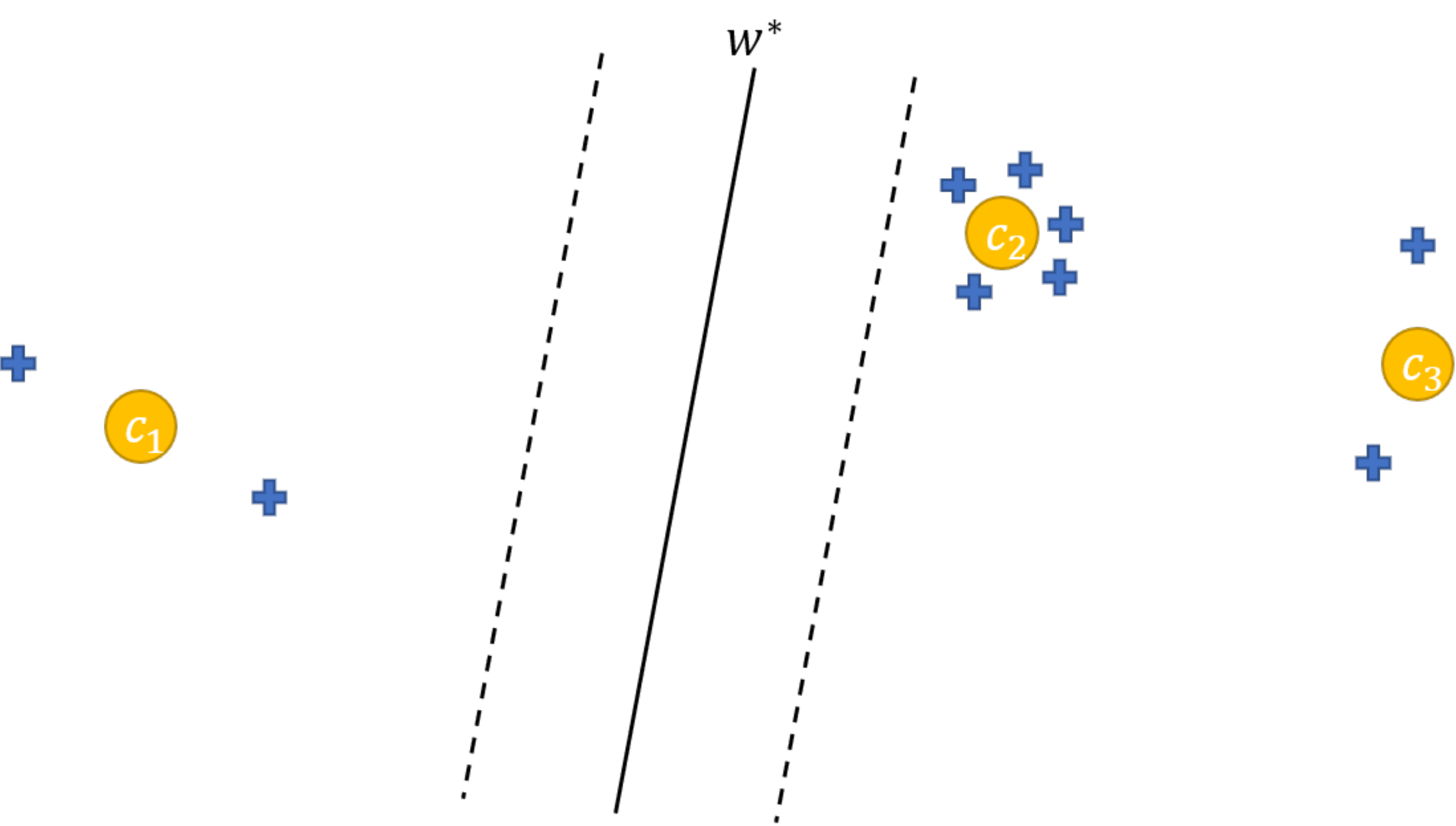}}
  \end{minipage}%

  \begin{minipage}[b]{1\textwidth}
  \centering
 \subcaptionbox{Points in small clusters have higher sensitivities, which quantifies the importance of outliers and misclassified points as they will be mostly in small clusters as $k$ goes larger.}[1\textwidth]{\includegraphics[width=1\textwidth]{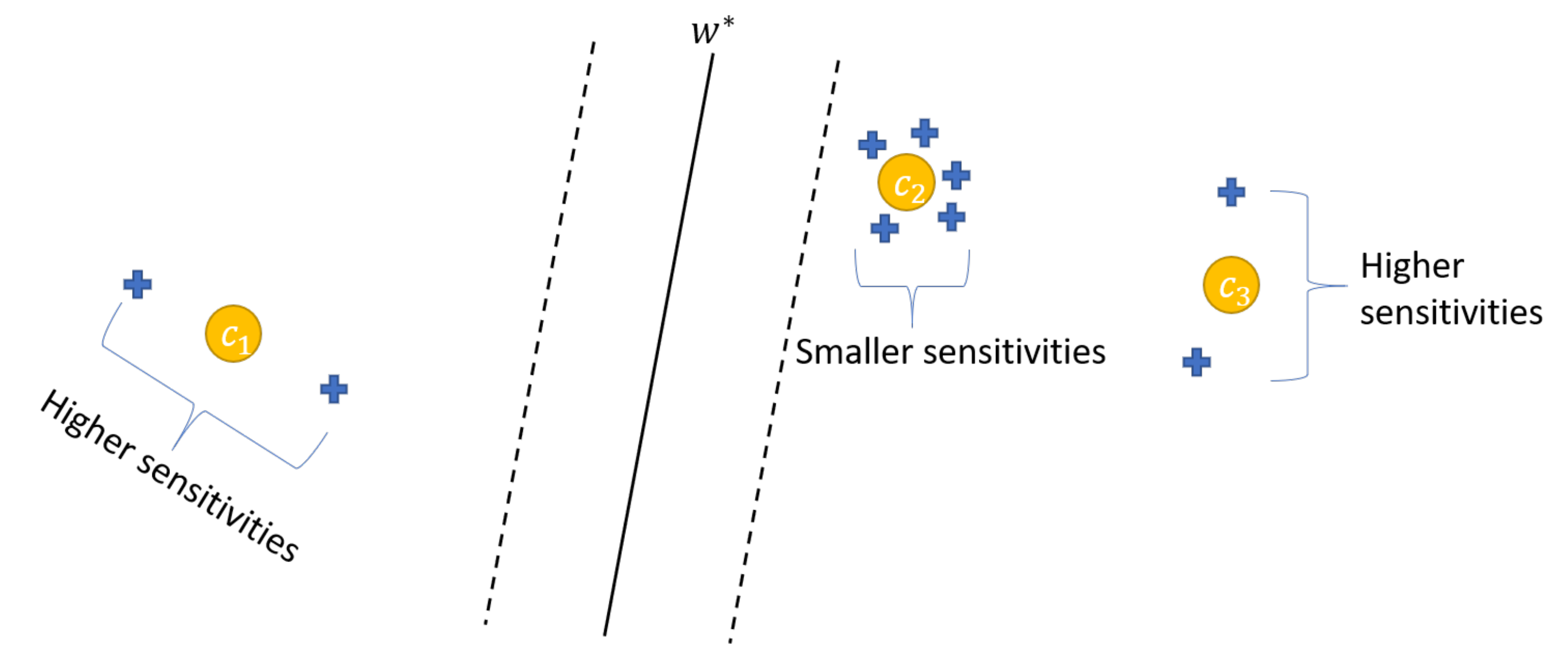}}
  \end{minipage}%
  \caption{Understanding the effect of $k$-means on the sensitivities of the points.}
  \label{fig:kmeans-effect}
\end{figure*}

In light of the above, we observe that as $k$ goes larger, the sensitivity of each point gets closer and closer to being simply the \emph{raw contribution}, which in case of unweighted data set, is simply applying uniform sampling. Thus, for each weighted $(P,u)$, we simply chose $k = \log{n}$ where $n$ denotes the total number of points in a $P$.

This observation helps in understanding how the sensitivities that we are providing for outliers and misclassified points actually quantifies the importance of such points. specifically speaking, as $k$ goes larger the outliers would mostly be assigned to the same cluster and since in general there aren't much of these points, we end up giving higher sensitivities for such points (than others) due to the fact that their \emph{raw contribution} increases as the size of the cluster, they belong to, decreases. When $k$ isn't large enough to separate these points from the rest of the data points, then the \emph{actual contribution} kicks into play, which then the mean of the cluster is shifted towards the ``middle'' between the outliers and the rest of the points in that cluster, boosting the \emph{actual contribution} of the rest of points inside the same cluster; See Fig.~\ref{fig:kmeans-effect}

\subsection{Towards finding the best $k$ value}
In the context of clustering, specifically speaking, $k$-means clustering problem, there is no definitive, provable way for determining the best $k$ value, while considering the computation cost needed for applying the $k$-means algorithm (or alternatively $k$-means++). However, there are some tools which can point to the best $k$ value from a heuristic's point of view. Such tools include the \emph{Silhouette Method} and the \emph{Elbow Method}. Fortunately enough, in our context, using the same methods can aid us in finding a ``good'' trade-off  between the \emph{raw contribution} and the \emph{actual contribution}.

Although, such methods are proven to be useful in practice~\cite{kodinariya2013review}, in our experiments, we simply chosen a constant value for $k$ depending on the number of points in the data set as elaborated in Section~\ref{sec:results}, and section~\ref{sec:exp-details}, which is shown to be useful in practice at Figure~\ref{fig:RelativeError} and Figure~\ref{fig:streaming}.

\section{Experimental Details}
\label{sec:exp-details}
Our experiments were implemented in Python and performed on a 3.2GHz i7-6900K (8 cores total) machine with 64GB RAM. We considered the following datasets in our evaluations.
\begin{enumerate}
  \item \textit{HTRU} --- $17,898$ radio emissions, each with $9$ features, of the Pulsar star.
  \item \textit{CreditCard} --- $30,000$ client entries each consisting of $24$ features that include education, age, and gender among other factors.
    \item \textit{Pathological} --- $1,000$ points in two dimensional space describing two clusters distant from each other of different labels, as well as two points of different labels which are close to each other.\footnote{We note that uniform sampling performs particularly poorly against this data set due to the presence of outliers.}
  \item \textit{Skin} --- $245,057$ random samples of B,G,R from face images consisting of 4 dimensions.
  \item \textit{Cod(-rna)} --- $488565$ RNA records consisting each of $8$ features.\footnote{This data set was attained by merging the training, validation and testing sets.}
  \item \textit{W1} --- $49,749$ records of web pages consisting each of $300$ features.
\end{enumerate}

\paragraph{Preprocessing step.}
Each data set has gone through a standardization process which aims to rescale the features so that they will have zero mean and unit standard deviation. As for the case where a data set is unweighted, we simply give each data point a weight of $1$, i.e., $u : P \to 1$ where $P$ denotes the data set, and the regularization parameter $\lambda$ was set to be $1$ throughout all of our experiments. 

\paragraph{$k$-means clustering.}
In our experiments, we set $k = \log{n}$ where $n$ is the number of points in the dataset (each datasets has different $k$ value). As for the clustering itself, we have applied $k$-means$++$~\cite{arthur2007k} on each of $\PP_+$ and $\PP_-$ as stated in our analysis; see Sec.~\ref{sec:analysis}.

\paragraph{Evaluation under streaming setting.} Under streaming setting, the range for sample sizes is the same as for running under offline settings (See Figures~\ref{fig:RelativeError} and~\ref{fig:timing}). What differs is the quality of the solver itself, which we use to show the effectiveness of our coreset compared to uniform sampling, i.e., we have chosen to make the solver (SVC of Sklearn) more accurate by lowering its optimal tolerance.

\section{Evaluations of Computational Cost for the Streaming Setting}
\label{sec:appendix-Timing}

\begin{figure*}[!htb]
  \centering
  \begin{minipage}[b]{0.32\textwidth}
  \centering
 \includegraphics[width=1\textwidth]{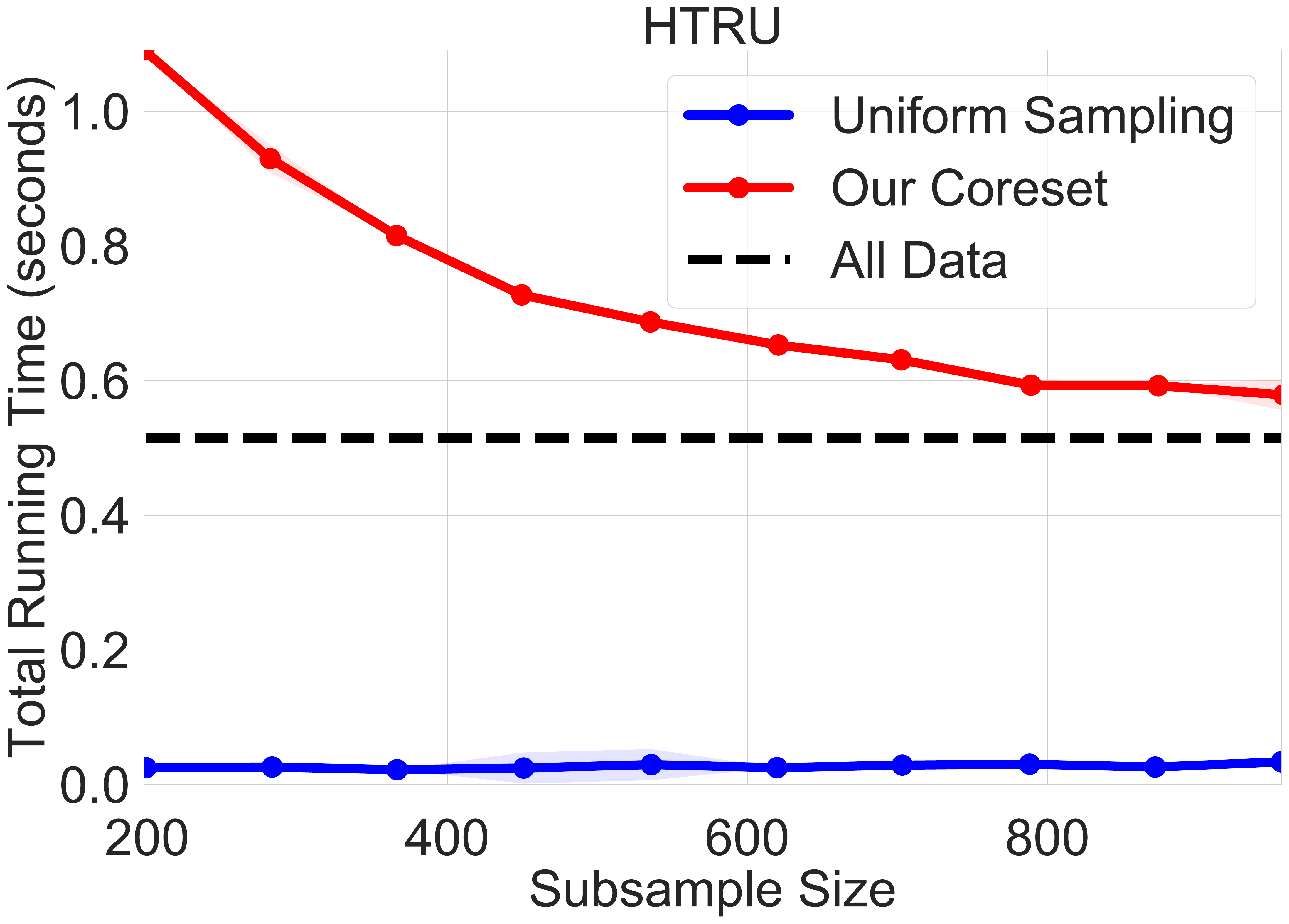}
  \end{minipage}%
  \begin{minipage}[b]{0.32\textwidth}
  \centering
 \includegraphics[width=1\textwidth]{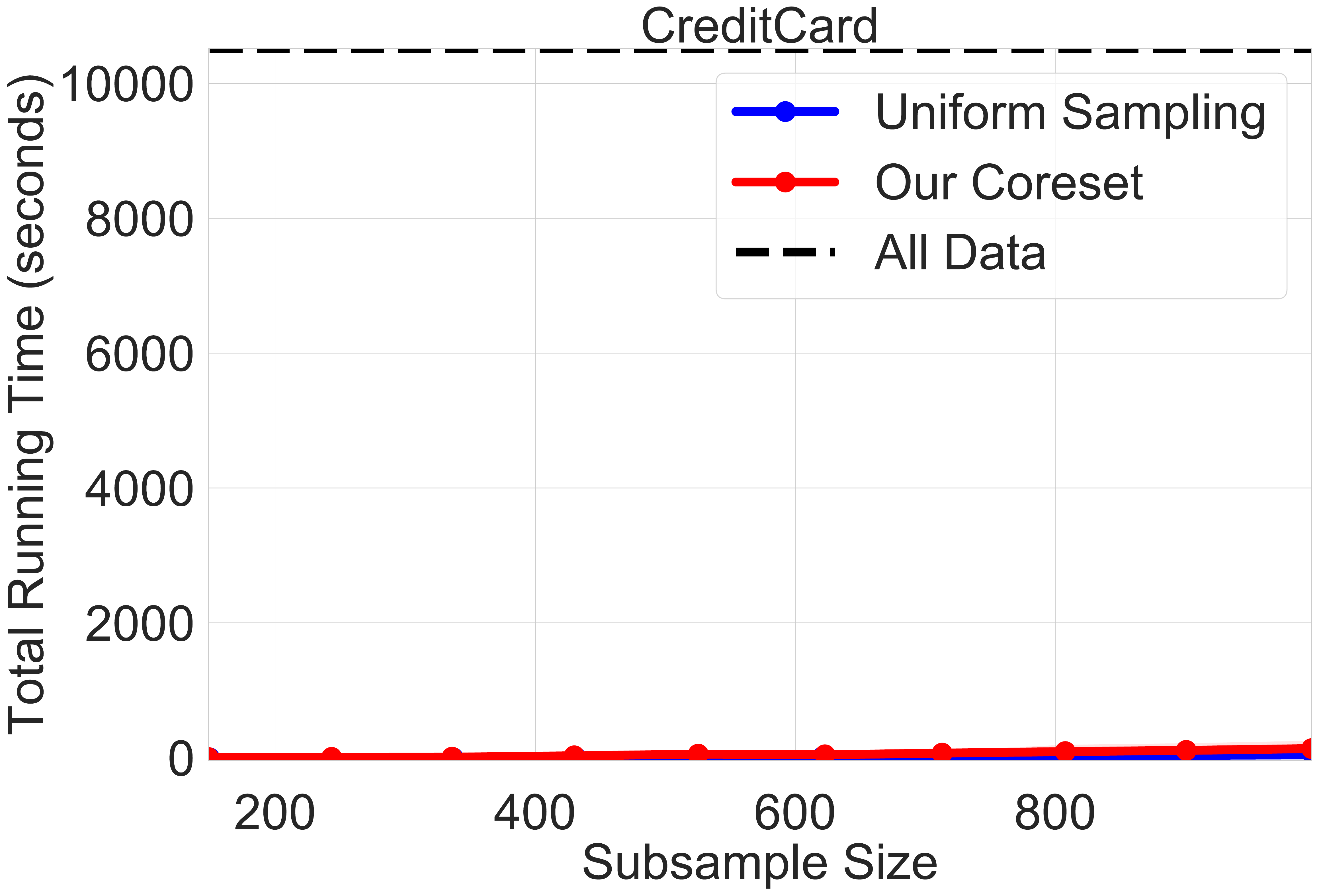}
  \end{minipage}%
  \begin{minipage}[b]{0.32\textwidth}
  \centering
 \includegraphics[width=1\textwidth]{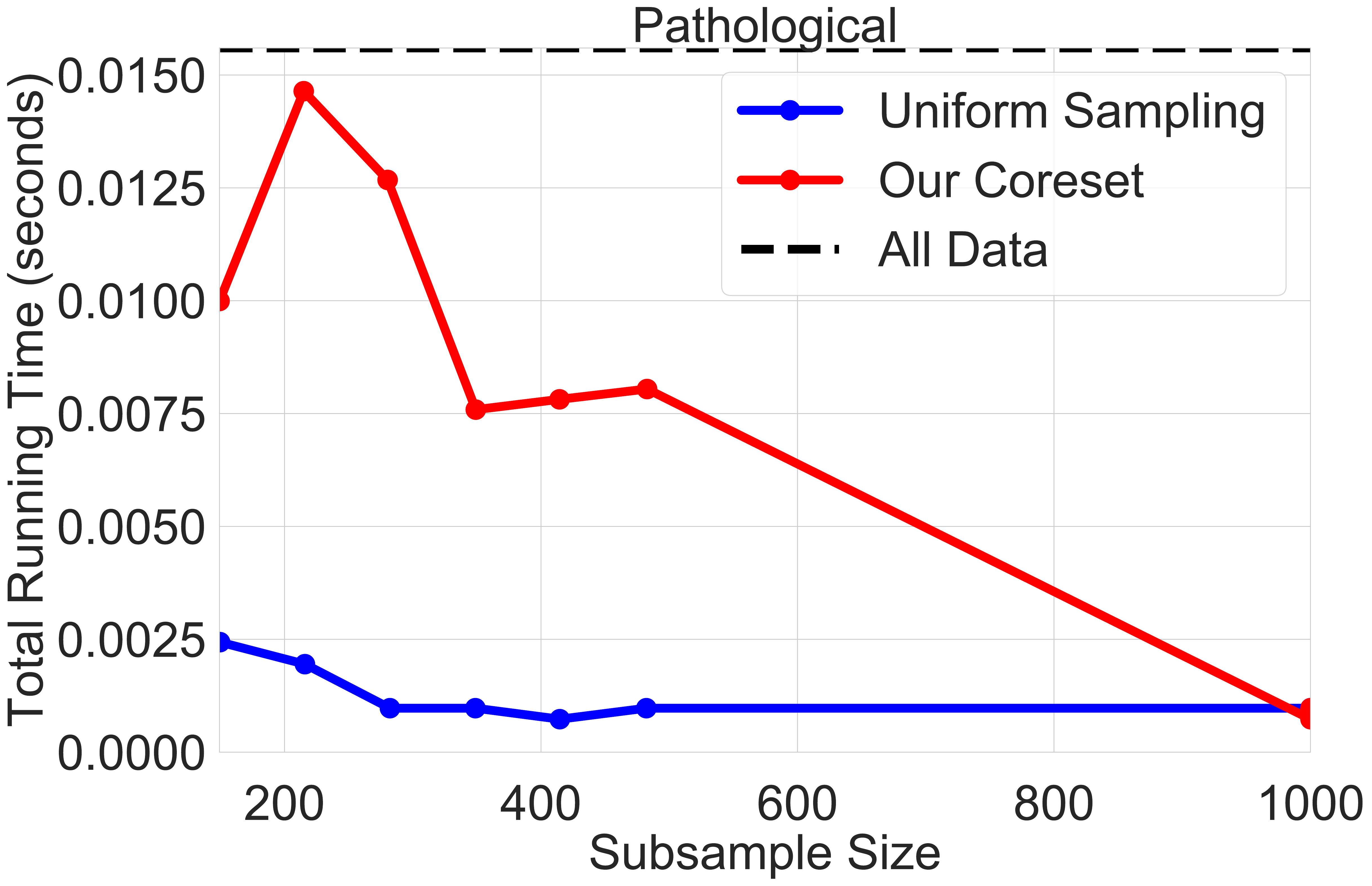}
  \end{minipage}%
  
\begin{minipage}[b]{0.32\textwidth}
  \centering
 \includegraphics[width=1\textwidth]{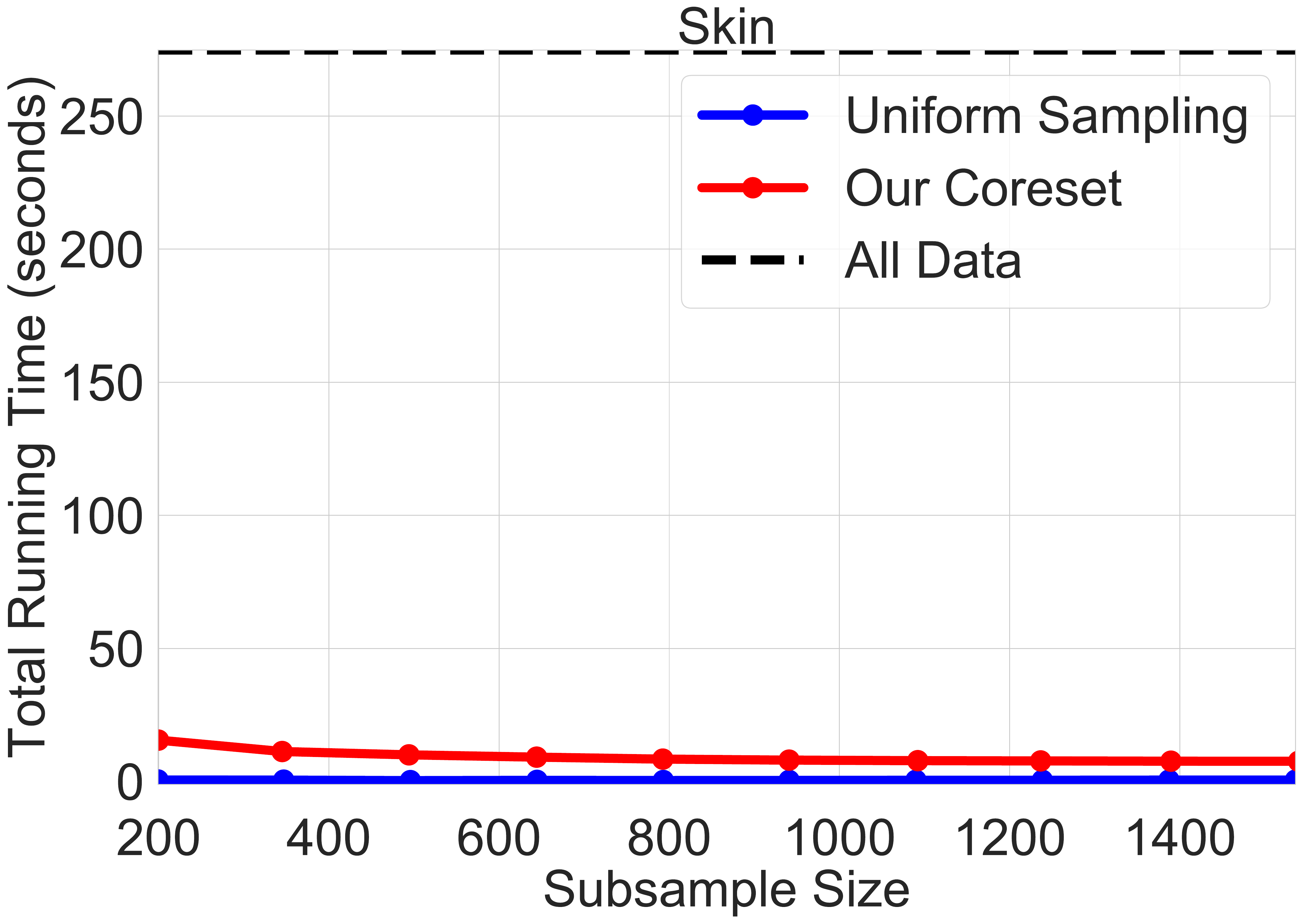}
  \end{minipage}%
  \begin{minipage}[b]{0.32\textwidth}
  \centering
 \includegraphics[width=1\textwidth]{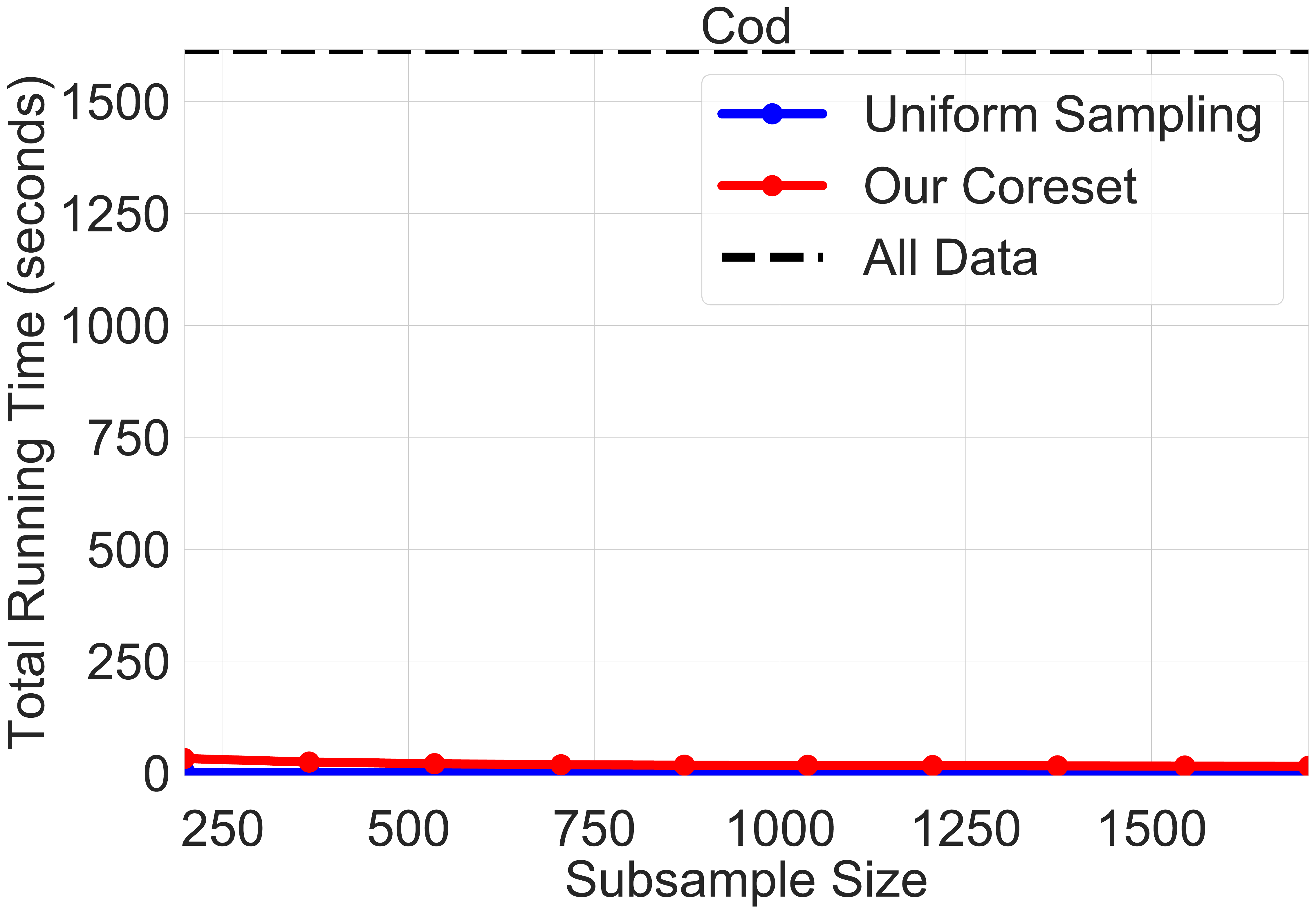} 
  \end{minipage}%
  \begin{minipage}[b]{0.32\textwidth}
  \centering
 \includegraphics[width=1\textwidth]{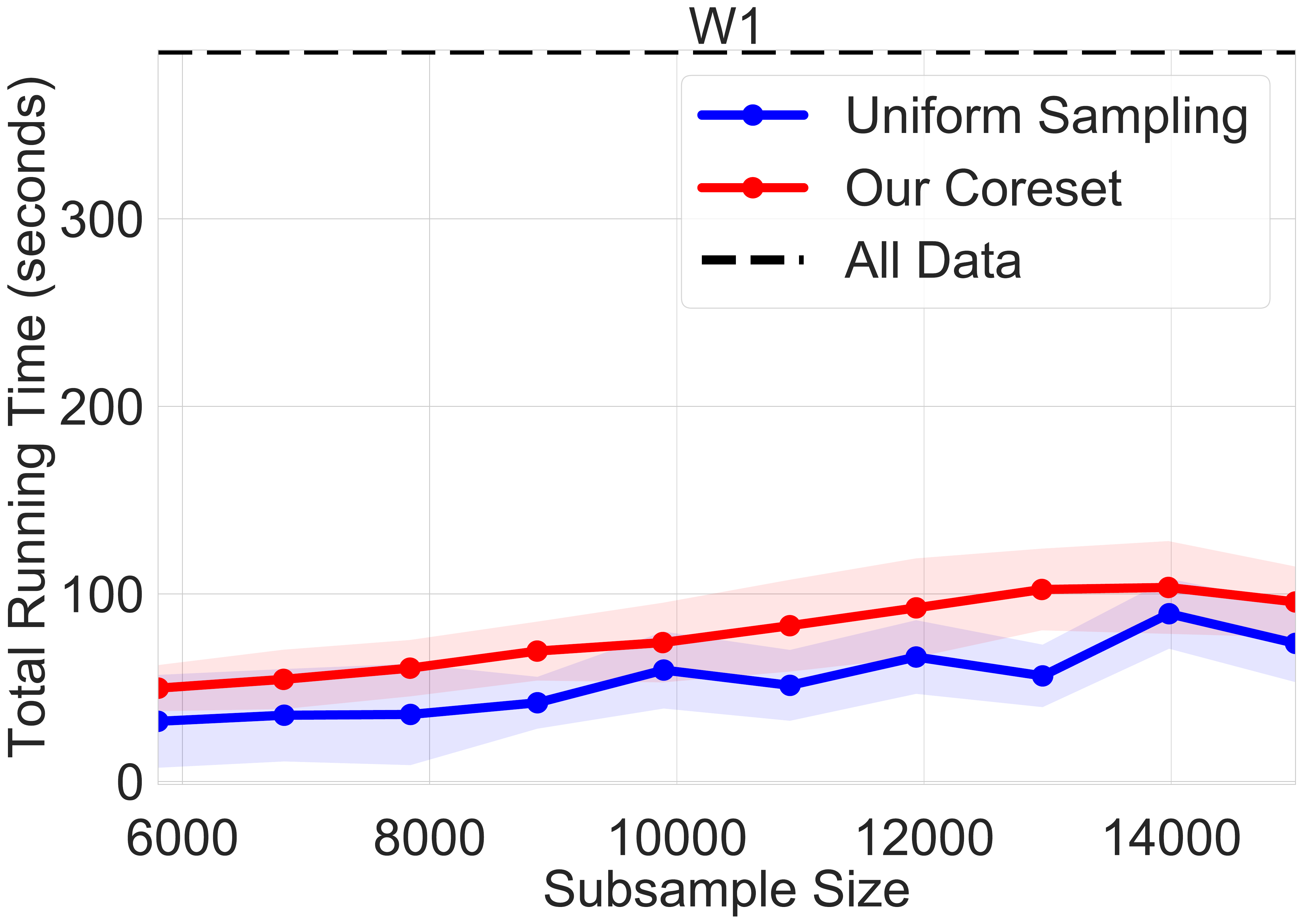}
  \end{minipage}%
    \caption{The \emph{total} computational cost of constructing a coreset using the merge-and-reduce tree and training the SVM model the coreset, plotted as a function of the size of the coreset.}
	\label{fig:streaming-timing}
\end{figure*}

\end{document}